\pdfoutput=1

\documentclass[11pt]{article}
\usepackage{url}
\usepackage[preprint]{acl}

\usepackage{times}
\usepackage{latexsym}

\usepackage[T1]{fontenc}

\usepackage[utf8]{inputenc}

\usepackage{microtype}

\usepackage{inconsolata}

\usepackage{graphicx}
\usepackage{subfigure}
\usepackage{algorithm,algorithmic}
\usepackage{float}          
\usepackage{multirow}   
\usepackage{booktabs}  

\usepackage{stfloats}     

\usepackage{pifont}        

\usepackage[skins]{tcolorbox}  
\tcbuselibrary{breakable}
\usepackage{boites, boites_exemples}
\usepackage{amsmath}
\usepackage{cleveref}
\usepackage{xspace}

\newcommand{\hb}{HammerBench\xspace}

\title{\hb: Fine-Grained Function-Calling Evaluation \\in Real Mobile Assistant Scenarios}

\author{
  Jun Wang$^{*,1}$ \hspace{0.5cm}
  Jiamu Zhou$^{*,1}$ \hspace{0.5cm}
  Muning Wen$^{2}$ \hspace{0.5cm}
  Xiaoyun Mo$^{1}$ \hspace{0.5cm}\\
  {\bf Haoyu Zhang$^{1}$ \hspace{0.5cm}
  Qiqiang Lin$^{1}$ \hspace{0.5cm}
  Cheng Jin$^{1}$ \hspace{0.5cm}
  Xihuai Wang$^{2}$}\\
  {\bf Weinan Zhang$^{2}$ \hspace{0.5cm}
  Qiuying Peng$^{+,1}$ \hspace{0.5cm}
  Jun Wang$^{+,1}$ \hspace{0.5cm}}\\
  \texttt{\{wangjun15,zhoujiamu,moxiaoyun,zhanghaoyu1,linqiqiang1,}\\
  \texttt{jincheng,pengqiuying,wangjun7\}@oppo.com} \\
  \texttt{\{muningwen,leoxhwang,wnzhang\}@sjtu.edu.cn}
}

\begin{document}
\maketitle

{\let\thefootnote\relax
\footnote{\footnotesize $^{*}$Equal Contribution, $^{+}$Corresponding Author, $^{1}$OPPO Research Institute, $^{2}$Shanghai Jiao Tong University}}


\begin{abstract}

Evaluating the performance of LLMs in multi-turn human-agent interactions presents significant challenges, particularly due to the complexity and variability of user behavior. In this paper, we introduce \hb, a novel benchmark framework for assessing LLMs' function-calling capabilities in real-world, multi-turn dialogues. \hb simulates diverse mobile assistant use cases, incorporating imperfect instructions, dynamic question-answer trajectories, intent and argument shifts, and the indirect use of external information through pronouns. To construct this benchmark, we curate a comprehensive dataset derived from popular mobile app functionalities and anonymized user logs, complemented by a cost-effective data generation pipeline leveraging open-source models. \hb is further augmented with fine-grained interaction snapshots and metrics, enabling detailed evaluation of function-calling performance across individual conversational turns. We demonstrate the effectiveness of \hb by evaluating several leading LLMs and uncovering key performance trends. Our experiments reveal that different types of parameter name errors are a significant source of failure across different interaction scenarios, highlighting critical areas for further improvement in LLM robustness for mobile assistant applications\footnote[1]{The code and datasets will be publicly available at \url{https://github.com/MadeAgents/HammerBench}.}.

\end{abstract}

\section{Introduction}
The mobile assistant built on large language models (LLMs), where users interact with agents to provide indispensable context for accurate API calling \citep{lin2024hammer}, often needs to handle multiple rounds of user interaction to complete complex task requests from users \citep{berkeley-function-calling-leaderboard}.
For instance, tasks like ticket purchasing typically demand multiple interactions to gather and provide complete context and information for accurate decisions. 
Such multi-turn interactions pose challenges for LLM-based agents in understanding user intent and making correct function calls due to the uncertainty of user intent and behavior.

To help detect the defects of LLM agents in personal assistant scenarios, many automated dataset evaluation methods driven by LLMs have been developed by the community including multi-turn dialogue \cite{ou2023dialogbench,bai2024mt} and function call evaluation\cite{berkeley-function-calling-leaderboard,NoisyToolBench}.
These function call benchmarks are typically built based on virtual environments to support quantifiable automated evaluation.
However, in complex multi-turn interactions, conversations and function calls are often closely intertwined \cite{MINT}, and current benchmarks struggle to evaluate this complex relationship.
And we still lack a fine-grained benchmark built in real-world scenarios data to identify the practical application flaws of the agent.

Based on the analysis of logs from anonymized real users, we found that in the real-world scenario, users may frequently change their intent during interactions, requiring the agent to accurately recognize these changes for successful task completion. Current benchmarks are unable to capture this process dynamic and lack the granularity needed to detect process defects or subtle issues in task execution. Additionally, in different scenarios, the references to external information and the differences in the information provided by users vary greatly, so the evaluation needs to cover a broader range of scenarios to help detect these differences.
Therefore, We argue that a comprehensive evaluation of mobile personal assistant agent has three key principles:

\textbf{Authenticity of test data} for capturing capabilities in satisfying the needs of real users.


\textbf{Diversity of scenarios} for capturing more situations that users may encounter in daily use.


\textbf{Granularity of evaluation} for capturing potential process defects.

In this paper, we introduces \hb, a fine-grained, multi-turn function-calling benchmark. 
To better reflect real-world user experiences, we curated the dataset based on popular mobile app functionalities from major app stores and queries derived from anonymized user logs. 
We expanded the dataset with open-source models, creating a cost-effective data generation pipeline. 
\Cref{table:comparison} shows \hb's superior diversity and coverage across various interaction scenarios. 
\hb supports a wider range of complexities in both single-turn and multi-turn dialogues, addressing key aspects often overlooked in previous benchmarks, such as handling imperfect responses, supporting diverse Q\&A trajectories, and dealing with indirect references or vague descriptions. 
\hb further introduces fine-grained metrics to evaluate function-calling performance across individual conversational snapshots. 
These included metrics such as Function Name Accuracy, Parameter Name Hallucination and Missing Rate, Progress Rate and Success Rate, offering a detailed assessment of the model's ability to handle function calls and track progress throughout multi-turn interactions. 

\begin{table*}[ht]
\centering
\footnotesize
\setlength\tabcolsep{3pt}
\begin{tabular}{ccccccccc}
\toprule
\multirow{4}{*}{Benchmark} & \multirow{4}{*}{\begin{tabular}[c]{@{}c@{}}Construction\\ Method\end{tabular}} & \multicolumn{3}{c}{Single-turn} & \multicolumn{3}{c}{Multi-turn} & \multirow{4}{*}{\begin{tabular}[c]{@{}c@{}}External\\ Individual\\ Information\end{tabular}} \\ \cmidrule(lr){3-5}\cmidrule(rl){6-8} & & Perfect & Imperfect & Irrelevant & Intent Shifts & \begin{tabular}[c]{@{}c@{}}Diverse Q\&A\\ Trajectories\end{tabular} & \begin{tabular}[c]{@{}c@{}}Argument\\ Shifts\end{tabular} & \\ 
\midrule
API-Bank & Human &\ding{56} &\ding{56} &\ding{56} &\ding{56} &\ding{56} &\ding{56} &\ding{56} \\
BFCL & LLM+Human &\ding{52} &\ding{56} &\ding{52} &\ding{52} &\ding{56} &\ding{56} &\ding{52} \\
NoisyToolBench & Human &\ding{56} &\ding{52} &\ding{52} &\ding{52} &\ding{56} &\ding{56} &\ding{56} \\
ToolSandBox & Human &\ding{52} &\ding{56} &\ding{52} &\ding{52} &\ding{56} &\ding{56} &\ding{56} \\ 
\midrule
\hb (Ours) & LLM+Human &\ding{52} &\ding{52} &\ding{52} &\ding{52} &\ding{52} &\ding{52} &\ding{52} \\ 
\bottomrule
\end{tabular}
\vspace{-0.5em}
\caption{Comparison between existing benchmarks and ours. \Cref{table:dataset description} provides detailed descriptions for each column.}
\label{table:comparison}
\vspace{-1.5em}
\end{table*}



To validate the effectiveness of \hb, we benchmark 10 LLMs with \hb and analyze their performance.
Our experiments demonstrate that \hb can perform more fine-grained evaluations in richer scenarios to assess the agent's performance in a broader range of real-world tasks. 
Specifically, \hb reveals significant challenges in multi-turn interactions, particularly in tracking argument shifts and handling external information, which affect the accuracy and consistency of function-calling performance across models.
Furthermore, \hb's fine-grained metrics, such as Progress Rate and Parameter Hallucination Rate, effectively highlight specific model issues, like hallucinations and parameter name errors, providing deeper insights into model performance across different contexts.
These results demonstrate that \hb can provide a comprehensive and in-depth evaluation for LLM agents in mobile assistant scenarios and offer valuable insights for optimizing robust function-calling LLMs.

\section{Related Works}
\subsection{Dialogue System Evaluation}


Evaluating dialogue systems is challenging due to the open-ended nature of human conversation, making traditional rule-based evaluations difficult. Early approaches, such as topic-based evaluations \citep{Topic-based-MT}, used topic classifiers to assess dialogue quality by evaluating sub-topics individually. With the rise of large language models (LLMs), newer methods leverage these models as evaluators. For example, \citep{zheng2023judging} discussed the advantages of using large models for dialogue assessment. Recent frameworks like MT-Bench \citep{bai2024mt} and MT-Eval \citep{kwan2024mteval} apply GPT-based models to evaluate multi-turn dialogues by decomposing them into sub-tasks. MT-Bench evaluates dimensions such as Context Memory, Anaphora Resolution, and Reasoning, while MT-Eval focuses on Recollection, Expansion, Refinement, and Follow-up. However, in function-calling scenarios, evaluating dialogue dynamics—rather than just the function call itself—adds a layer of complexity, requiring a nuanced approach to assessment.

\subsection{Function Calling Benchmark}

Function-calling benchmarks can be broadly categorized into single-turn and multi-turn evaluations. Single-turn benchmarks like ToolAlpaca \citep{ToolAlpaca}, ToolLLM \citep{ToolLLM}, and NexusRaven \citep{nexusraven} use automatic data generation based on LLM world knowledge. In contrast, benchmarks such as ToolBench \citep{ToolBench} and RestGPT \citep{restGPT} rely on manually labeled data to ensure quality. Some benchmarks, including BFCL-V1 and SealTool \citep{sealtool}, use the self-instruct method \citep{wang-etal-2023-self-instruct} to generate data, while BFCL-V2 refines datasets by cleaning online user logs. Despite these advancements, single-turn benchmarks struggle to capture the diversity of real-world user interactions and fail to fully assess function-calling capabilities.

Multi-turn function-calling benchmarks, such as API-Bank \citep{li-etal-2023-api} and NoisyToolBench \citep{NoisyToolBench}, are manually curated, ensuring high-quality dialogue data. MINT \citep{MINT} focuses on interactive tools and external tools in question answering, while BFCL-V3 and ToolSandBox \citep{berkeley-function-calling-leaderboard, lu2024toolsandbox} offer comprehensive multi-turn evaluations. However, gaps remain in these benchmarks, especially regarding argument shifts.

In the broader landscape of agent evaluations, multi-turn frameworks like AgentBoard \citep{ma2024agentboard} and Agent as Judge \citep{zhuge2024agent} provide general evaluation systems for agent capabilities. However, these systems do not specifically target the challenges of multi-turn function-calling scenarios. To address the limitations of existing benchmarks, we propose a more targeted evaluation system that focuses on the fine-grained aspects of multi-turn function calling, including argument shifts and external information handling.

\section{Principles of HammerBench}

\begin{table*}[htb]
\centering
\scriptsize
\setlength\tabcolsep{4pt}
\resizebox{1\textwidth}{!}{
\begin{tabular}{clcl}
\toprule
Evaluation Type                                                                               & \multicolumn{1}{c}{Data Type} & Abbr. & \multicolumn{1}{c}{Description} \\ \midrule
\multirow{4}{*}{Single-turn}                                                                            & Single-turn perfect instruction & Perfect & The user query that clearly gives all required parameter values. \\
& Single-turn imperfect instruction & Imperfect & The user query that only gives few required parameter values. \\ 
& Single-turn with pronouns & External & There are anaphoric pronouns that refer to external individual information in the user query. \\ 
& Single-turn irrelevant & Irrelevant & There is no tool in the candidate tools list that can solve the user query. \\ 
\midrule
\multirow{4}{*}{\begin{tabular}[c]{@{}c@{}}Diverse \\question-answer \\ trajectories\end{tabular}} & Single-question-single-answer & sQsA & The agent asks and the user answers with parameter values one by one. \\
& Multi-question-multi-answer & mQmA & The agent asks for multiple parameter values, the user also answers the corresponding value. \\ 
& Multi-question-single-answer & mQsA & The agent asks for multiple parameter values, but the user only answers one value. \\ 
& Single-question-multi-answer & sQmA & The agent only asks for one parameter value, but the user answers multiple values. \\ 
\midrule
Intent shifts  & User intent has changed & IS & \begin{tabular}[c]{@{}l@{}}The agent should output some special tags to terminate the ongoing session when a user\\  expresses a new intent unrelated to the current slot filling process, allowing for the \\ subsequent handling of the new intent.\end{tabular} \\ 
\midrule
\multirow{2}{*}{\begin{tabular}[c]{@{}c@{}}Argument shifts\end{tabular}} & Slots overriding & SO & The user repeatedly modifies the value of the same slot before tool execution. \\ 
& Multiple slot values & mSv & The user repeatedly modifies the value of the same slot after tool execution. \\ 
\midrule
\begin{tabular}[c]{@{}c@{}}External \\ individual \\ information\end{tabular} & Answering with pronouns & External & The user does not answers directly, but gives the external individual information pronoun.\\ 
\bottomrule
\end{tabular}
}
\vspace{-0.5em}
\caption{Descriptions of test data in \hb at different granularities.}
\label{table:dataset description}
\vspace{-2em}
\end{table*}

Based on the analysis of anonymized user logs and existing benchmarks, we identify three key principles for designing an ideal benchmark: authenticity, diversity, and granularity. Below, we discuss how \hb incorporates these principles.

\textbf{Authenticity of test data: }
Queries should reflect real user behavior. However, user logs often exhibit a long-tail distribution, where rare but impactful scenarios may be underrepresented. To ensure comprehensive evaluation, we supplement the dataset with additional instances from these rare scenarios. Moreover, traditional API designs prioritize developer-centric parameters, which may not be intuitive for end users. \hb prioritizes user-friendly tools, abstracting API complexities to maintain operational integrity while providing a seamless experience. The benchmark dataset is curated from real Apps' functionalities and anonymized user logs, capturing a wide range of common and rare user intents.

\textbf{Diversity of scenarios: }
A diverse range of tools, queries, and user behaviors is essential for assessing LLM agents’ function-calling abilities. \hb tests agents across multiple domains through a diverse range of tools, from ticket buying to daily schedules. It also includes various query types, from simple requests to complex tasks, and captures diverse user behaviors, such as detailed vs. vague queries and single-turn vs. multi-turn interactions. This diversity enables the benchmark to evaluate how well the agent adapts to different contexts and user expectations. In short, our benchmark encompasses four categories in multi-turn interactive tasks: 1) Imperfect instructions (e.g., unclear or vague queries), 2) Varied question-answer trajectories (e.g., differing conversational paths), 3) Intent and argument shifts (e.g., changing goals or inputs), and 4) External individual information (e.g., implicit user context or background).

\textbf{Granularity of evaluation: }
Granularity refers to the level of detail in the evaluation metrics used to assess system performance. \hb adopts fine-grained metrics to evaluate both individual function-calling tasks and multi-turn interactions. Key metrics include Function Name Accuracy, Parameter Name Hallucination and Missing Rate, and Argument Accuracy. These metrics enable a detailed assessment of specific performance aspects, such as correct function invocation and handling of incomplete information. In addition, \hb tracks the Progress Rate, which measures how well the model progresses toward task completion, and the Success Rate, which evaluates the final outcome. This granular approach allows us to identify subtle deficiencies in model behavior, such as mismanagement of dependencies or errors in reasoning, providing valuable insights for further optimization.

\section{HammerBench}





In this section, we describe the methodology and workflow for constructing the datasets used in \hb.
To provide a more concise overview, the overall structure of test data generated for \hb is summarized in \Cref{table:dataset description}, with detailed statistics and analysis in \Cref{sec:analysis}.

\begin{figure*}[htb] 
    \vspace{-1em}
    \centering 
    \includegraphics[width=1\textwidth]{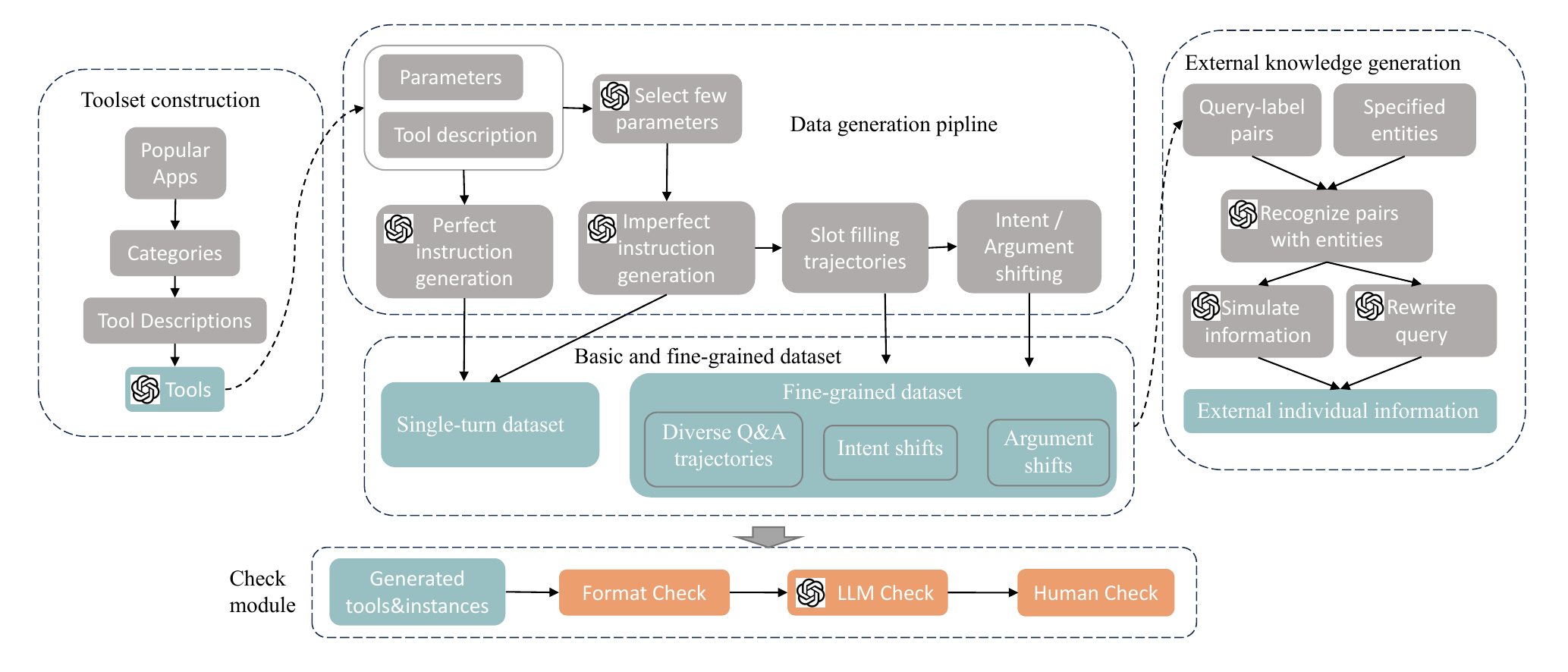} 
    \vspace{-2.5em}
    \caption{\hb construction pipeline: toolset collection, data generation, external knowledge generation, and validation. Blocks with GPT icons indicate the use of LLMs, while orange blocks represent verification modules, and green blocks denote various data types corresponding to each phase.} 
    \label{fig:generation pipline} 
    \vspace{-1em}
\end{figure*}

\subsection{Test Data Construction}

\hb's dataset construction follows a four-stage systematic process: toolset collection, API generation, validation, and manual refinement, as shown in \Cref{fig:generation pipline}.

\textbf{Tool Collection and API Generation:} The toolset is based on functionalities from popular mobile apps sourced from major manufacturers' app stores. We categorize these apps into 60 functional groups, for which we create prompts (see \Cref{appx:Tool Generation}) to generate APIs using LLMs. This process involves creating function names, descriptions, parameters, and return values, with human oversight to ensure simplicity, user-friendliness, and minimal redundancy.

\textbf{Validation and Refinement:} Generated APIs undergo manual review for quality, resulting in 1,063 final selections. The review process prioritizes completeness (covering typical usage scenarios) and rationality (clarity and simplicity without losing functionality). Function-calling instances are generated using the self-instruct method (\citealp{wang-etal-2023-self-instruct}), producing three types of instructions: (1) complete function calls, (2) incomplete calls (missing key parameters), and (3) irrelevant calls (triggering rejection). This ensures both practicality and usability for effective evaluation.

\subsubsection{Multi-Turn Interactions with Imperfect Instructions}\label{sec:imperfect}

As 76\% of queries in user logs contain fewer than 10 tokens, we designed a workflow to generate datasets that reflect typical query lengths. First, we use LLMs to identify the most commonly used parameters in daily applications, with the prompt for parameter selection provided in \Cref{appx:Parameters Selection}. LLMs then generate queries including these parameters (see \Cref{appx:Query Generation}). If parameters are missing or extraneous, the model regenerates the query until it aligns with the preset list. 

\begin{figure*}[htb] 
    \vspace{-2em}
    \centering 
    \includegraphics[width=1\textwidth]{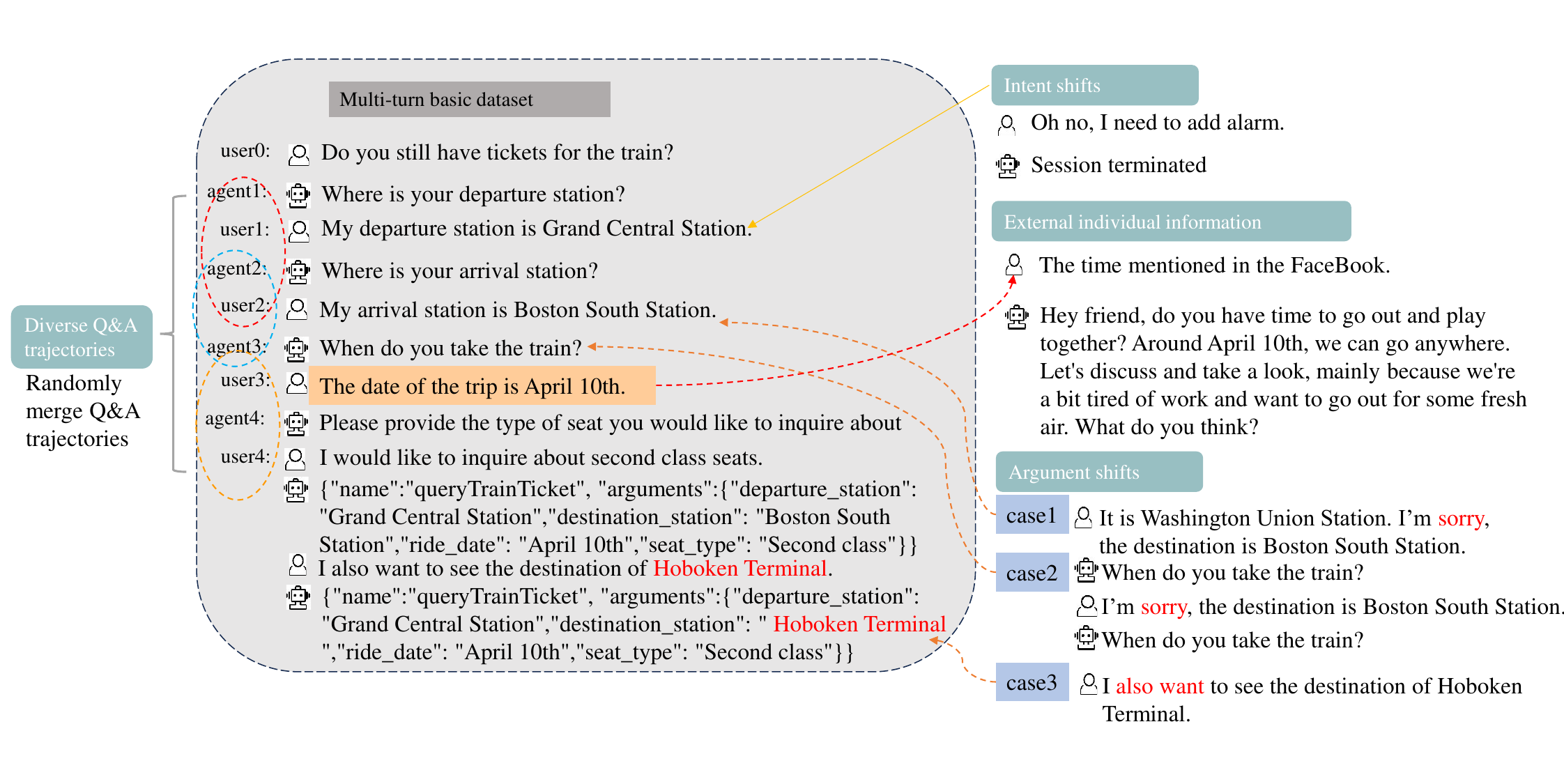} 
    \vspace{-3em}
    \caption{Examples of four types of test cases in \hb: 1) Diverse Q\&A trajectories generated by merging user-agent interactions; 2) Intent shifts: agent terminates the session when users change their intent; 3) Argument shifts: three cases of changing slot values during interactions; 4) External individual information: users use pronouns instead of exact details, common in real-world interactions.} 
    \label{fig:examples of dataset} 
    \vspace{-1em}
\end{figure*}

This process produces imperfect instances that pass basic format verification, but hallucinations remain a concern. To mitigate this, we introduce a semantic validation step (\citealp{li-etal-2024-one}), checking if the function call generated by the LLM is consistent across zero-shot and one-shot in-context settings. Consistency is evaluated with two metrics: 1) Rouge-L, using a predefined threshold, and 2) semantic alignment, by re-assessing the function call with the LLM. Instructions that are consistent in zero-shot settings are deemed more reliable, reducing hallucination risk. Instances that pass validation are further refined through manual inspection and sampling review. This results in 3,240 imperfect instances, forming the basis for the multi-turn function-calling scenarios discussed in the following sections, as exampled in \Cref{fig:examples of dataset}.

\subsubsection{Diverse Question-Answer Trajectories}\label{sec:qa}

Building on the imperfect instructions from the previous section, we extend these to multi-turn function-calling dialogues with interactive slot-filling. To assess the model’s ability to handle diverse conversational flows, we design four types of question-answer trajectories:

\textbf{Single-Question-Single-Answer:} The model queries a single argument, and the user provides a corresponding value. This tests the model’s ability to accurately elicit a piece of information. 

\textbf{Multi-Question-Multi-Answer:} The model asks multiple questions, and the user answers each. This evaluates the model’s ability to process multiple inputs within a conversation. 

\textbf{Single-Question-Multi-Answer:} The model asks one question, and the user gives multiple responses. This challenges the model to interpret multiple answers for a single argument. 

\textbf{Multi-Question-Single-Answer:} The model asks multiple questions, but the user only answers one. This scenario tests the model’s ability to handle incomplete information and adapt to unexpected behavior.

We generate these trajectories by first creating single-question-single-answer instances using the prompts in \Cref{appx:Multi-turn Prompt}, which undergo semantic and manual review (\Cref{sec:imperfect}). The remaining trajectories are formed by randomly combining questions and answers from this dataset, resulting in 2,310 distinct instances for evaluation.






\subsubsection{Intent and Argument Shifts}

In real-world interactions, users often shift their intentions or modify slot values unexpectedly. For example, users may abandon a complex request or switch goals mid-conversation. This section evaluates whether the model can recognize intent shifts and respond accordingly, such as by issuing a rejection signal to terminate the conversation. To create the dataset for intent shifts, we begin with \textit{Single-Question-Single-Answer} instances. We then randomly replace a user input with a function call for a different API, simulating a sudden intent change, resulting in 1,098 intent shift instances. 

Additionally, we consider two types of argument shifts: slot overriding and API repurposing.

\textbf{Slot Overriding:} Users may provide different values for the same parameter, with only the most recent one being relevant. We generate these instances by modifying dialogues to include turns where users explicitly override previous arguments, resulting in 1,462 instances.

\textbf{API Repurposing:} Users may update certain parameters of a previous function call while keeping others unchanged. To model this, we add a user input turn after a function call, introducing new values for some parameters while retaining others, yielding 1,066 instances.





\subsubsection{References to External Individual Data}

In AI assistant interactions, users often refer to external data indirectly using pronouns, instead of explicitly stating specific values. To evaluate the model's anaphora resolution ability, we propose a method to generate datasets with references to external data. These references can come from any function call instance, with data retrieved from various external sources.

We select nine common information entities (e.g., names, company names, product IDs, license plates, etc.) and use LLMs to identify argument values corresponding to these entities. Instances containing such references are retained for further processing. One identified argument is then replaced by a pronoun referring to the external entity, and a new instance is generated by rephrasing the original text to include the pronoun. All instances are validated semantically and manually to ensure accuracy. This process results in 1,175 single-turn and 487 multi-turn instances involving external references.

\begin{table*}[th]
\centering
\small
\setlength\tabcolsep{4pt}
\resizebox{\textwidth}{!}{
\begin{tabular}{lccccccccc}
\toprule
\multirow{4}{*}{Model}  & \multirow{4}{*}{Overall (Acc.)} & \multicolumn{4}{c}{Single-turn (Acc.)}   & \multicolumn{4}{c}{Multi-turn (Acc.)} \\ 
\cmidrule(lr){3-6} \cmidrule(lr){7-10} 
                                    &   & Perfect & Imperfect & External & Irrelevant & \begin{tabular}[c]{@{}c@{}}Diverse\\Q\&A\end{tabular} & \begin{tabular}[c]{@{}c@{}}Argument\\shifts\end{tabular} & External & \begin{tabular}[c]{@{}c@{}}Intent\\shifts\end{tabular}  \\ \midrule

GPT-4o (Prompt)                     & \underline{71.39\%}        & 78.69\% & \underline{78.77\%}   & 56.16\%  & 69.52\%    & \underline{73.80\%}      & \underline{68.34\%}         & \underline{69.64\%}  & 76.22\%       \\
Claude3.5-sonnet (Prompt)           & 70.28\%       & \underline{79.86\%} & \textbf{79.38\%}   & 55.83\%  & 68.16\%    & \textbf{74.85\%}      & \textbf{73.62\%}         & \textbf{71.83\%}  & 58.74\%       \\
\midrule
Qwen2.5-72B-Instruct (Prompt)       & \textbf{71.50\%}      & \textbf{80.86\%} & 68.64\%   & \textbf{67.40\%}  & \underline{73.09\%}    & 65.93\%      & 65.22\%         & 61.12\%  & \underline{89.79\%}       \\
Qwen2.5-7B-Instruct (Prompt)        & 58.80\%     & 75.75\% & 58.08\%   & 58.89\%  & 41.00\%    & 57.57\%      & 53.21\%         & 53.87\%  & 72.04\%       \\
Llama-3.1-70B-Instruct (Prompt)     & 61.58\%     & 75.99\% & 69.47\%   & \underline{59.57\%}  & 12.49\%    & 73.13\%      & 69.67\%         & 69.29\%  & 63.02\%       \\
Llama-3.1-8B-Instruct (Prompt)      & 49.09\%      & 73.01\% & 51.08\%   & 57.61\%  & 8.13\%     & 53.51\%      & 48.61\%         & 50.14\%  & 50.63\%       \\
Ministral-8B-Instruct (Prompt) & 37.42\%      & 78.02\% & 32.19\%   & 57.70\%  & 3.86\%     & 39.15\%      & 38.68\%         & 35.00\%  & 14.75\%       \\
\midrule
Hammer2.1-7b (FC)         & 65.79\%   & 68.85\% & 59.29\%   & 50.21\%  & \textbf{89.74\%}    & 59.10\%      & 55.73\%         & 48.73\%  & \textbf{94.71\%}       \\
xLAM-7b-fc-r (FC)         & 50.79\%     & 73.48\% & 48.82\%   & 48.85\%  & 50.02\%    & 52.30\%      & 46.20\%         & 48.30\%  & 38.34\%       \\
ToolACE-8B (FC)          & 50.52\%     & 78.82\% & 51.94\%   & 56.93\%  & 26.58\%    & 49.35\%      & 48.35\%         & 49.15\%  & 42.98\%       \\ 
\bottomrule
\end{tabular}
}
\caption{\hb leaderboard (Single-Turn and Multi-Turn), where the best performances are highlighted.}
\label{table:leaderboard}
\vspace{-1.5em}
\end{table*}

\subsection{Function Calling Snapshots}

In complex interactions, models must assess whether the historical context provides enough information to fulfill user intent and identify missing details for API parameters. When information is incomplete, agents should generate follow-up questions to request the missing inputs. A common approach is the Learning to Ask paradigm, where models are fine-tuned to generate questions at each turn and then make a function call once all parameters are gathered.

We suggest an alternative, the Function Calling Snapshots, where models generate a formatted function call, or Snapshot, at each turn, regardless of context completeness. This call is then compared to the function definition using rule-based logic to identify missing information, enabling the generation of relevant follow-up questions. By decoupling the identification of missing data from question generation, this approach allows for more focused information recognition and detailed analysis, such as detecting Missing Parameters or Hallucination.

We evaluated this mechanism through a comparative experiment by fine-tuning Qwen2.5-7B on both paradigms and testing on 100 out-of-distribution (OOD) conversations with human evaluation. As shown in \Cref{table:snapshot}, the Function Calling Snapshots approach outperformed Learning to Ask by 16\% in task success rate, demonstrating its superior effectiveness.

\begin{table}[htb]
\centering
\begin{tabular}{ll}
\toprule
Training Paradigm  & SR \\
\midrule
Learning to Ask    & 68\%  \\
Function Calling Snapshots & 84\%  \\
\bottomrule
\end{tabular}
\caption{Comparison between Learning to Ask and Function Calling Snapshots paradigms.}
\label{table:snapshot}
\vspace{-1em}
\end{table}

\subsection{Fine-Grained Evaluation Metrics}

We use several key metrics to assess function-calling performance:

\textbf{Accuracy (Acc.):} a traditional metric, the accuracy when function and parameter names are all correctly predicted.

\textbf{Function Name Accuracy (Func. Acc.):} the accuracy of predicted function names, without taking the parameters into account.

\textbf{Function-Irrelevant Accuracy (Irrelevant):} the proportion of samples where irrelevant functions are correctly excluded.

\textbf{Parameter Hallucination and Missing Rate (PHR/PMR):} PHR tracks incorrect parameter names, while PMR tracks missing parameters. These measure the accuracy and completeness of parameter predictions.

\textbf{Progress Rate (PR):} the proportion of correct function calls up to the turns of error, defined as $PR = \frac{k}{n}$, where $k$ is the number of correct turns and $n$ is the total number of turns.

\textbf{Success Rate (SR):} the overall accuracy of function calls across all turns in a conversation, with higher values indicating better final performance.

These metrics—Acc., Func. Acc., PHR, PMR, SR, and PR—provide a comprehensive framework for evaluating function calls in both individual and multi-turn dialogues. 


\section{Experiment}




In this section, we demonstrate the superiority of \hb through a series of experiments with 10 commonly used or function-calling-specific LLMs. 
We provide detailed analyses of the experimental results, followed by in-depth discussions and valuable insights that could be used to guide subsequent model optimization.

\subsection{Experimental Setups}

To validate the effectiveness of \hb, we benchmark the following baseline models: GPT-4o (\citealp{hurst2024gpt}), Claude3.5-sonnet(\citealp{anthropic2024introducing}), Llama-3.1-70B-Instruct (\citealp{dubey2024llama}), Qwen2.5-72B-Instruct (\citealp{qwen2.5}), Llama-3.1-8B-Instruct, Qwen2.5-7B-Instruct, Ministral-8B-Instruct (\citealp{mistral}), xLAM-7b-fc-r (\citealp{liuapigen}) and ToolACE-8B (\citealp{liu2024toolace}). For all models, we impose a strict requirement on the output format in the prompt, as detailed in \Cref{appx:function calling prompt}. All open-source models are deployed on 4 A100 80G GPUs, with a temperature setting of 0.0 and bfloat16 precision.

\begin{table*}[th]
\centering
\small
\setlength\tabcolsep{4pt}
\resizebox{\textwidth}{!}{
\begin{tabular}{cccccccccccc}
\toprule
\multirow{3}{*}{Model}   & \multicolumn{4}{c}{Diverse Q\&A (PR)}      & \multicolumn{4}{c}{Argument shifts (PR)}   & \multicolumn{2}{c}{External (PR)} &  \multirow{3}{*}{IS (Acc.)}                \\ \cmidrule(lr){2-5} \cmidrule(lr){6-9} \cmidrule(lr){10-11}
                                    & sQsA    & mQmA    & mQsA    & sQmA    & w/o SO  & w/ SO   & w/o mSv & w/ mSv  & w/o External  & w/ External &                     \\ \midrule
GPT-4o (Prompt)                     & \underline{66.03\%} & \underline{73.22\%} & \underline{68.52\%} & \underline{70.79\%} & \underline{67.90\%} & \underline{68.65\%} & \underline{68.47\%} & 56.45\% & \underline{74.36\%}       & \underline{66.34\%}    & 76.22\%             \\
Claude3.5-sonnet (Prompt)           & 70.64\% & 73.74\% & 68.93\% & 72.62\% & 71.71\% & 72.54\% & 71.99\% & 63.94\% & 76.60\%       & 68.62\%  & 58.74\%             \\
Qwen2.5-72B-Instruct (Prompt)       & 56.28\% & 64.75\% & 56.93\% & 64.19\% & 61.11\% & 60.25\% & 63.15\% & 57.35\% & 65.94\%       & 59.58\%  & \underline{89.79\%}             \\
Llama-3.1-70B-Instruct (Prompt)     & 64.72\% & 67.42\% & 61.63\% & 68.00\% & 65.34\% & 62.01\% & 65.70\% & \underline{58.64\%} & 72.55\%       & 63.47\%     & 63.02\%             \\
Qwen2.5-7B-Instruct (Prompt)        & 48.90\% & 53.06\% & 43.62\% & 53.04\% & 50.66\% & 45.78\% & 51.71\% & 41.75\% & 58.13\%       & 50.04\%   & 72.04\%             \\
Llama-3.1-8B-Instruct (Prompt)      & 38.43\% & 51.15\% & 38.42\% & 50.35\% & 41.54\% & 37.99\% & 42.76\% & 34.97\% & 47.80\%       & 43.56\%   & 50.63\%             \\
Ministral-8B-Instruct-2410 (Prompt) & 26.78\% & 28.59\% & 19.02\% & 28.37\% & 27.76\% & 26.21\% & 28.17\% & 22.00\% & 31.34\%       & 25.06\%   & 14.75\%             \\
Hammer2.1-7b (FC)                   & 52.37\% & 57.57\% & 41.50\% & 57.18\% & 53.87\% & 49.83\% & 54.76\% & 43.87\% & 60.62\%       & 46.32\%   & 94.71\%             \\
xLAM-7b-fc-r (FC)                   & 39.28\% & 40.74\% & 30.36\% & 41.08\% & 39.01\% & 34.44\% & 39.14\% & 29.09\% & 49.07\%       & 39.76\%   & 38.34\%             \\
ToolACE-8B (FC)                     & 42.55\% & 46.39\% & 25.87\% & 46.10\% & 44.08\% & 40.07\% & 44.81\% & 35.07\% & 51.12\%       & 43.61\%   & 42.98\%             \\ \bottomrule
\end{tabular}
}
\caption{Ablation results: The evaluation (PR) on \hb for different multi-turn data types, while ``w/o'' refers to subsets of sQsA datasets with the same user queries with ``w/''. The IS represents the success rate of LLM terminating correctly when encountering intent shifts. This table is a snippet from \Cref{table:Overall}.}
\label{table:PR}
\vspace{-1.5em}
\end{table*}

\begin{table*}[htb]
\centering
\tiny
\setlength\tabcolsep{5pt}
\resizebox{1.\textwidth}{!}{
\begin{tabular}{cccccccc}
\toprule
\multirow{3}{*}{Model}              & \multirow{3}{*}{Func. Acc.} & \multicolumn{3}{c}{First snapshot}                      & \multicolumn{3}{c}{Last snapshot}                        \\ \cmidrule(lr){3-5} \cmidrule(lr){6-8} 
                                    &                           & PHR  & PMR & Other Error Rate & PHR  & PMR  & Other Error Rate\\ 
                                    \midrule
GPT-4o (Prompt)                     & 88.80\%                   & 8.51\%  & 3.38\% & 2.87\%                 & 6.56\%  & 8.82\%  & 8.41\%                    \\
Claude3.5-sonnet (Prompt)           & 90.16\%                   & 8.59\%  & 2.42\% & 1.52\%               & 6.97\%  & 7.98\%  & 4.75\%                   \\
Qwen2.5-72B-Instruct (Prompt)       & 85.43\%                   & 12.79\% & 1.28\% & 1.81\%              & 10.98\% & 5.97\%  & 4.05\%                 \\
Llama-3.1-70B-Instruct (Prompt)     & 91.62\%                   & 19.09\% & 2.19\% & 1.99\%                & 11.13\% & 2.09\%  & 5.86\%                 \\
Qwen2.5-7B-Instruct (Prompt)        & 82.88\%                   & 24.73\% & 2.86\% & 1.32\%               & 14.62\% & 5.82\%  & 5.49\%               \\
Llama-3.1-8B-Instruct (Prompt)      & 89.25\%                   & 36.84\% & 1.63\% & 1.22\%              & 15.20\% & 22.14\% & 4.18\%               \\
Ministral-8B-Instruct-2410 (Prompt) & 74.59\%                   & 52.50\% & 2.93\% & 0.98\%              & 18.07\% & 6.96\%  & 5.37\%                 \\
Hammer2.1-7b (FC)                   & 83.06\%                   & 19.08\% & 3.73\% & 1.21\%                & 13.27\% & 9.43\%  & 4.82\%                  \\
xLAM-7b-fc-r (FC)                   & 88.80\%                   & 47.18\% & 3.38\% & 1.03\%             & 18.05\% & 8.72\%  & 4.10\%                \\
ToolACE-8B (FC)                     & 81.79\%                   & 29.62\% & 3.90\% & 1.22\%               & 25.50\% & 4.90\%  & 4.45\%            \\ \bottomrule
\end{tabular}
}
\caption{Evaluations on the first and last snapshots in the sQsA dataset. First snapshot: the first turn of the conversation, Last snapshot: the last turn of the conversation. Other Error Rate: parameter values error rate.}
\label{table:MT first last}
\vspace{-2em}
\end{table*}

\subsection{Results Analysis of Evaluation Types}


\Cref{table:leaderboard} shows the overall result of \hb in different evaluation types including single-turn and multi-turn.
Models like Qwen2.5-72B-Instruct (Prompt), GPT-4o (Prompt), and Claude3.5-sonnet (Prompt) are among the top performers, with a strong ability to handle both single-turn and multi-turn interactions, while models like Hammer2.1-7b (FC) and Ministral-8B-Instruct (Prompt) show more specialized strengths or weaknesses depending on the task type. Specifically, with the help of diverse scenarios, the following key insights primarily emerge regarding Intent Shifts, Argument Shifts, and External Individual Information.


\textbf{Intent Shifts.} In the scenario of intent shifts, we assess the LLM's ability to recognize changing user intentions. As shown in \Cref{table:leaderboard}, LLMs with stronger single-turn irrelevance detection capabilities tend to perform better in handling intent shifts.

\textbf{Argument Shifts.} As seen in experiments from \Cref{table:PR}, a major challenge in multi-turn interactions occurs when users modify the same slot multiple times before or after function execution. This tests the model's ability to track and update slot values. LLMs often struggle to handle such shifts, like when a user changes a date from June 1st to July 8th but the model retains the old value. This issue is particularly evident in scenarios with similar parameters, such as updating train orders. The Slot Overriding task effectively detects this flaw, revealing that many open-source LLMs are slow to adapt to changes, while GPT-4o shows stronger performance in tracking dynamic slot updates. Another challenge arises when users modify a slot after a tool has been executed, leading to issues with maintaining consistency across multiple values. Even GPT-4o, though less affected, sometimes fails to track all updated parameters. This is likely due to the model perceiving the tool call as “complete” after execution, which causes it to disregard prior input. The Multiple Slot Values task highlights this issue, where slot values are inconsistently managed across turns. This task underscores the difficulty models face in maintaining parameter consistency over extended interactions. \Cref{appx:badcase argument shifts} shows examples of these challenges.

\textbf{External Information.} User queries containing external individual information often introduce noise, affecting slot-filling accuracy. The External Information task effectively detects this issue, showing how LLMs struggle with indirect references like pronouns or contextual information, leading to lower accuracy as shown in \Cref{table:PR} and Appendix \ref{appx:badcase MTEK}. This task reveals the vulnerability of LLMs to external factors like personalized data or context, significantly disrupting function-calling accuracy. It highlights the importance of models being able to separate core inputs from external noise, making it a critical metric for real-world performance. Further results are available in \Cref{appx: more res}.

\subsection{Results Analysis of Evaluation Metrics}

In addition to the results from diverse scenarios, the fine-grained metric design of \hb allows for a more detailed identification of specific issues in models within particular contexts. We will analyze the effectiveness of the metrics from different scenarios. Some interesting insights are revealed with the metrics defined in \hb, with more details in \Cref{appx: more res}.

\textbf{Progress Rate. }
In diverse Q\&A dataset \Cref{table:PR}, interactions often lead open-source LLMs to hallucinate additional parameters, particularly when addressing multi-question-single-answer (mQsA) scenarios, which results in parameter mispredictions, as elaborated in \Cref{appx:badcase random merge agent}. Progress Rate (PR) can help gauge model resilience effectively. For instance, GPT-4o demonstrates minimal degradation in PR, suggesting its robustness against hallucination issues.




\textbf{PHR/PMR.} Based on the results presented in \Cref{table:MT first last}, the PHR in the initial snapshot, where the context is incomplete, is significantly higher compared to the final snapshot, where the context is complete. This indicates that an incomplete context leads to a higher frequency of hallucinations in parameters, which remains a persistent bottleneck for many open-source LLMs.

\section{Conclusion}



This paper presents \hb, a fine-grained evaluation benchmark for multi-turn function calling tasks, assessing imperfect instructions, diverse question-answer trajectories, intent and argument shifts, and user queries with external individual information. These components reflect real-world user scenarios. We decompose multi-turn slot-filling interactions into a series of function-calling snapshots, allowing evaluation of each conversational turn. Additionally, we introduce random merging of dialogue trajectories to assess question-answer trajectory variability. A method for generating data with external individual information is also released, enabling evaluation of a model's ability to resolve pronouns and anaphora. 

Our experimental results highlight the challenges LLMs face in handling argument shifts and external information, e.g. tracking dynamic slot updates. Additionally, metrics such as Progress Rate and PHR/PMR reveal critical areas of improvement, notably in model resilience against hallucinations and handling evolving user intents, underscoring the need for further advancements to enhance LLM robustness in real-world applications.

\section*{Limitations}


Although \hb can assess certain multi-turn real-world scenarios and facilitates the evaluation of multi-turn interactions in more complex contexts by incorporating multi-dimensional data generation methods, its evaluations are confined to fixed, pre-generated datasets. This limitation inherently restricts the benchmark's capacity to fully capture the diverse range of behaviors and capabilities that a large language model (LLM) might exhibit in a dynamic, real-world setting. When compared to the approach of using LLMs to generate real-time, dynamic trajectories during the evaluation process(\citealp{lu2024toolsandbox}), the scope of trajectories covered by \hb is less comprehensive, and it does not assess the model's ability to self-explore. 

The datasets used in \hb are automatically generated by synthesizing real-world scenarios, which inevitably introduces some discrepancies when compared to actual user queries. While this benchmark does support the evaluation of multi-turn interactions in more complex scenarios by incorporating data generation techniques from various dimensions, exhaustively capturing all possible user queries remains infeasible. The reliance on fixed evaluation trajectories, while making the evaluation process more convenient and controllable, limits \hb to only a partial step towards automated evaluations that more closely mirror real user interactions.

\bibliography{custom}

\appendix

\onecolumn

\section{Analysis of Benchmark Dataset}
\label{sec:analysis}
The overall structure of the datasets generated for \hb is summarized in \Cref{table:dataset description}. The dataset comprises four single-turn datasets, each with four distinct types, and eight multi-turn datasets, categorized into three types.

\begin{table}[htb]
\centering
\small
\begin{tabular}{cccc}
\toprule
\textbf{Statistics} & \ Perfect & Imperfect & External \\ 
\midrule
\# of categories    & 60                                                     & 60                                                       & 55                                                      \\
\# of tools         & 1063                                                   & 894                                                      & 463                                                     \\
\# of queries       & 2116                                                   & 3240                                                     & 1175                                                    \\ \bottomrule
\end{tabular}
\caption{Statistics of single-turn datasets.}
\label{table:statistics}
\end{table}

\begin{figure*}[htb]
  \centering
  \subfigure[Distribution  of \#parameter]{
  \label{Fig.sub.1}
  \includegraphics[width=0.31\textwidth]{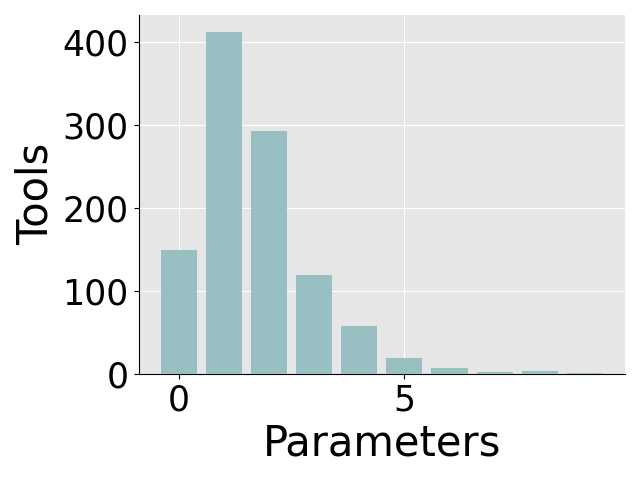}
  }
  \subfigure[Distribution of conversation turns]{
  \label{Fig.sub.2}
  \includegraphics[width=0.31\textwidth]{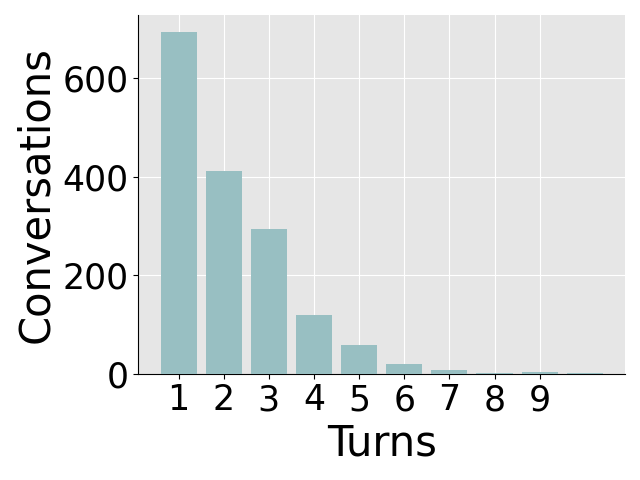}}
  \subfigure[Distribution of data types]{
  \label{Fig.sub.3}
  \includegraphics[width=0.31\textwidth]{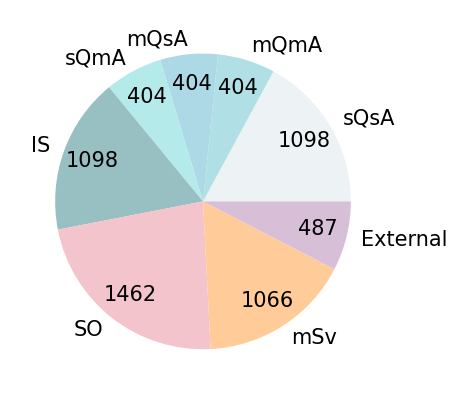}}
  \caption{Statistics: a)The number of tools corresponding to different parameter counts in our toolset; b)The number of conversations corresponding to different turn counts in sQsA; c)Various number of multi-turn data cases constructed based on the Imperfect and External in \Cref{table:statistics}.}
\end{figure*}

\textbf{Statistics:} Our \hb dataset includes 60 functional categories, 1,063 tools, and a total of 6,531 queries, distributed across three types, as shown in \Cref{table:statistics}. \Cref{Fig.sub.1} illustrates the correlation between the number of tools and the number of parameters defined in the toolset. For the parts of multi-turn test data, there are 1098 sQsA conversations generated from single-turn imperfect instructions, spanning 59 categories and 494 APIs. If we define the number of turns as the total number of questions asked by the agent, there are approximately 404 conversations with turns greater than 1. \Cref{Fig.sub.2} presents the distribution of conversations according to the number of turns. Because conversations can only be merged with more than one turn, there are 404 conversations in mQmA, mQsA, and sQmA. And the distribution of multi-turn dataset types is depicted in \Cref{Fig.sub.3}.

\textbf{Quality:} The dataset quality is rigorously ensured through the LLM validation module. Initially, data that fails the LLM check is manually corrected on a case-by-case basis. Data that passes the LLM's double-check are subject to a random sample review, with 100 instances selected for human recheck. Our quality assurance procedures have shown that the data availability rate exceeds 95\%. Given that our fine-grained multi-turn dataset is constructed from single-question-single-answer multi-turn data, the quality of the underlying single-turn instances is of paramount importance. Consequently, we conducted a manual review of 1,098 multi-turn instances to ensure their compliance with the logical flow of everyday conversations.

\newpage

\section{Examples}
\subsection{Single-turn}
\label{appx:single turn examples}
\begin{boiteepaisseavecuntitre}{normal user query}
\noindent \textbf{User:} Please help me send the traffic violation record of PDX-1234 in Portland yesterday to Manager Zhang by email.\\
\noindent \textbf{Agent:} \{"name":"shareViolationDetail", "arguments":\{"city": "Portland", "media": "email", "name": "Manager Zhang", "plate\_number": "PDX-1234", "time":"yesterday"\}\}
\end{boiteepaisseavecuntitre}

\begin{boiteepaisseavecuntitre}{missing arguments user query}
\noindent \textbf{User:} Share the traffic violation information in Portland.\\
\noindent \textbf{Agent:} \{"name":"shareViolationDetail", "arguments":\{"city": "Portland", "media": "", "name": "", "plate\_number": "", "time":""\}\}
\end{boiteepaisseavecuntitre}

\begin{boiteepaisseavecuntitre}{user query with external individual information}
\noindent \textbf{User:} I found the license plate number mentioned in my notepad and would like to share the traffic violation information I encountered in Portland yesterday with Manager Zhang via email. Is that okay?\\
\noindent \textbf{EK:} At 3pm today, I saw a car with license plate number PDX-1234 in the parking lot. The owner seemed to have forgotten to turn off the lights, and the lights were on until the evening. I am worried about violating traffic rules. If you know the owner, please remind him.\\
\noindent \textbf{Agent:} \{"name":"shareViolationDetail", "arguments":\{"city": "Portland", "media": "email", "name": "Manager Zhang", "plate\_number": "PDX-1234", "time":"yesterday"\}\}
\end{boiteepaisseavecuntitre}

\subsection{Multi Turn}
\label{appx:multi turn examples}
\begin{boiteepaisseavecuntitre}{conversation with external individual information}
\noindent \textbf{User:} Share the violation information.\\
\noindent \textbf{Agent:} Which city did the violation occur in?\\
\noindent \textbf{User:} The violation occurred in Portland.\\
\noindent \textbf{Agent:} Please provide a way to share.\\
\noindent \textbf{User:} Share the link via email.\\
\noindent \textbf{Agent:} Please provide the recipients of the shared content.\\
\noindent \textbf{User:} The recipient was Manager Zhang.\\
\noindent \textbf{Agent:} Please provide the license plate number.\\
\noindent \textbf{User:} The license plate number mentioned in the notebook\\
\noindent \textbf{EK:} At 3pm today, I saw a car with license plate number PDX-1234 in the parking lot. The owner seemed to have forgotten to turn off the lights, and the lights were on until the evening. I am worried about violating traffic rules. If you know the owner, please remind him.\\
\noindent \textbf{Agent:} When is the order opening time?\\
\noindent \textbf{User:} It was yesterday.\\
\noindent \textbf{Agent:} \{"name":"shareViolationDetail", "arguments":\{"city": "Portland", "media": "email", "name": "Manager Zhang", "plate\_number": "PDX-1234", "time":"yesterday"\}\}
\end{boiteepaisseavecuntitre}

\newpage

\section{Examples of Model Output Errors}
\subsection{Imperfect instruction}
\label{appx:badcase param hallucination}
\begin{boiteepaisseavecuntitre}{Parameter hallucination}
\noindent \textbf{User:} Help me find a mail.\\
\noindent \textbf{Agent:} \{"name":"searchMail", "arguments":\{"keyword": "mail", "max\_results": "10"\}\}
\end{boiteepaisseavecuntitre}
The user does not provide any parameter information, but the parameter slots are filled in by LLMs.

\subsection{Diverse Q\&A trajectories}
\label{appx:badcase random merge agent}
\begin{boiteepaisseavecuntitre}{Parameter hallucination}
\noindent \textbf{User:} See how much airline points I have.\\
\noindent \textbf{Agent:} What's your rider's name Provide the rider's ID number.\\
\noindent \textbf{User:} The passenger's name is Bob.\\
\noindent \textbf{Agent:} \{"name":"getFlightAirlinePoints", "arguments":\{"name": "Bob", "passenger\_id": "123456789"\}\}
\end{boiteepaisseavecuntitre}
Due to aske ID number, although the user does not answer, LLMs also mistakenly fills in this parameter slot.

\subsection{External Individual Information}
\label{appx:badcase MTEK}
\begin{boiteepaisseavecuntitre}{Redundant Information}
\noindent \textbf{User:} Share my taxi ride.\\
\noindent \textbf{Agent:} What time is the order\\
\noindent \textbf{User:} The time mentioned in the Facebook message sent to me by Xiao Ming<EK>:Yesterday evening at 8pm, we had a virtual meeting to discuss next week's work schedule. Everyone put forward a lot of good suggestions and felt that the meeting was efficient. Hopefully our team can continue to maintain this positive attitude!\\
\noindent \textbf{Agent:} \{"name":"shareTaxiOrder", "arguments":\{"time": "yesterday evening at 8pm", "media": "Facebook"\}\}
\end{boiteepaisseavecuntitre}
The user only mentions that the time is in the Facebook, but does not mention that the media is Facebook.

\subsection{Argument shifts}
\label{appx:badcase argument shifts}
\begin{boiteepaisseavecuntitre}{slot overriding}
\noindent \textbf{User:} Is there any way to change the train ticket standby order I applied for previously?\\
\noindent \textbf{Agent:} What number of trips did you book\\
\noindent \textbf{User:} The number I booked is G123. I'm sorry, Booked trips is D789.\\
\noindent \textbf{Agent:} \{"name":"updateTrainWaitlistOrder", "arguments":\{"train\_number": "D789", "after\_number": "G123"\}\}
\end{boiteepaisseavecuntitre}
The user provides two values for the same slot, but LLMs fail to recognize.

\begin{boiteepaisseavecuntitre}{multiple slot values}
\noindent \textbf{User:} Find me a way. How do I get there?\\
\noindent \textbf{Agent:} Please provide your departure location name or address If using your current location please confirm directly\\
\noindent \textbf{User:} People's Square\\
\noindent \textbf{Agent:} Please provide your destination name or address\\
\noindent \textbf{User:} Shanghai Hongqiao International Airport\\
\noindent \textbf{Agent:} \{"name":"vehicleNavigationRoute", "arguments":\{"departure": "People's Square", "destination": "Shanghai Hongqiao International Airport"\}\}\\
\noindent \textbf{User:} I also want to see that Name or address of the destination is Bondi Beach\\
\noindent \textbf{Agent:} \{"name":"vehicleNavigationRoute", "arguments":\{"destination": "Bondi Beach"\}\}
\end{boiteepaisseavecuntitre}
When another destination is given again, the original departure parameter is lost.

\newpage

\section{Prompts of Function Calling}
\label{appx:function calling prompt}
\begin{boiteepaisseavecuntitre}{format prompt}
\noindent You have access to the following functions.\\<tools>\\
\noindent To call a function, please respond with JSON for a function call.Respond in the format ```json\{"name": function name, "parameters": dictionary of argument name and its value\}.```\\
\noindent Attention! For time parameters, please fill in the format as described in the user request, and do not automatically convert the format! For example:\\user: Set an alarm for 8 a.m. tomorrow\\assistant: \{"name":"UtilityTools.AlarmClock.addAlarm", "arguments":\{"time":"8 a.m. tomorrow"\}\}\\\\\#\# Note! Please do not hallucinate parameters. If some parameters are not mentioned in the user request, please directly output an empty string '' For example:\\user:Check my traffic violation record to see what happened.\\assistant:```json\{"name": "Navigation.TrafficViolations.viewViolationDetail", "arguments":\{"plate\_number": "","city": "","time": ""\}\}```.\\Never ask to the user for missing parameters!Output tool call!\\\#\# If all the above tools are not suitable, you must output: Sorry, no tool is suitable for your request.\\\\Let's start!
\end{boiteepaisseavecuntitre}

\section{Prompts of Dataset Construction}

\subsection{Tool Generation}
\label{appx:Tool Generation}
\begin{boiteepaisseavecuntitre}{Generating Tools}
\noindent \# System\\You are a tool builder! Your task is to generate realistic and versatile toolkits that will be used by Large Language Models (LLMs) to enhance their ability to solve real-world tasks. You must generate toolkits that are useful, cohesive, complete, and diverse. You should focus on generating toolkits that are commonly used by average users.

\noindent \#\#\# Core requirements\\1. Realism requirement: Ensure that the generated toolkit has a realistic design and can be seamlessly integrated with LLMs for practical real-world applications. Specifically, the toolkit should either have a tangible prototype that has been deployed in the real world, such as those developed using existing APIs (such as Weibo or NetEase Mail), or demonstrate strong potential for future development and accessibility through APIs (such as industrial automation toolkits).\\2. Diversity requirement: Ensure that the generated toolkit is diverse, meeting a wide range of domains, types, functions, and use cases.\\3. Compatibility requirement: Ensure that the generated toolkit is compatible with the textual interfaces of LLMs. In particular, LLMs can only interact with the toolkit through textual and programmatic APIs. Therefore, the tool API should mainly accept and return text as input and output. Therefore, the toolkit's API should mainly accept and return text as input and output. In cases where the input or output involves alternative types such as files, images, videos, or audio, these should be interfaced through data paths or URLs, rather than raw data.\\4. Cohesion requirement: Ensure that the generated toolkit is a cohesive collection of related tools designed to facilitate the completion of a specific core target task. It should contain a variety of tool APIs that serve different purposes within the scope of the intended functionality. For example, the NetEase Mailbox toolkit may include tools for sending, searching, reading, and deleting emails.\\5. Completeness requirement: Ensure that the toolkit is functionally complete and self-contained. Each toolkit should contain comprehensive tool APIs that are sufficient to complete its core target tasks without relying on any external tools or resources. In particular, if a toolkit involves operating data resources (such as tasks in the Trello toolkit), the tool APIs should generally support relevant CRUD (create, read, update, delete) operations on these resources, or similar operations. In addition, these tool APIs should generally operate using unique identifiers of the data source. You should not assume that such unique identifiers will be provided by the user. Instead, there should be a tool API (e.g., a search tool API) in the toolkit for retrieving unique identifiers. A typical design of a complete toolkit is as follows:\\- A search tool for retrieving unique identifiers (and possibly primary content) of data resources, such as by keyword search.\\- A read tool that takes a unique identifier as a parameter and returns the detailed data resource.\\- An update tool that takes a unique identifier and the updated data resource as parameters and updates the data resource, returning the unique identifier of the updated data resource.\\- A delete tool that takes a unique identifier as a parameter and deletes the data resource, returning the success status of the deletion.\\- A create tool that takes a data resource as a parameter and creates a new data resource, returning the unique identifier of the created data resource.

\noindent \#\#\# Additional Requests\\1. Unique target user requirement: Ensure that the toolkit is designed for a specific target user group and that all tools will be used by the same target user group. The designated user group can vary - individual users, enterprise users, developers, etc. For example, in the case of an e-commerce toolkit, all tools should target either individual users or enterprise users, but not both.\\2. Compactness requirement: Ensure that the toolkit API is compact, meaning that if tools have similar interfaces and functionality, they should be unified into one tool. For example, tools for searching for different types of items should be unified into one tool API with a parameter for specifying the type of items to search for.\\3. Search limit requirement: For toolkits that involve searching the tool API to retrieve data resources, ensure that there is a parameter for specifying the maximum number of results returned by the search tool API. This parameter can be optional, and the default value is specified in the tool description.\\4. Avoid unnecessary authentication requirements: In most cases, toolkits are designed for personal use by a single user, and it is reasonable to assume that the user has completed authentication before using the toolkit. In this case, ensure that all tools do not require login, authentication, or personal ID. For example, all tools in a personal banking toolkit should not require a login or a `user\_id` parameter for the tool. This also means that "unauthorized evaluation" should not be considered a potential risk for the toolkit.\\5. Avoid unnecessary ID requirements: Ensure that the tool does not introduce unnecessary unique identifiers. Unique identifiers are only necessary when multiple data resource instances of the same type need to be distinguished and operated on by the unique identifier. For example, for an online shopping toolkit, an `order\_id` needs to be introduced because there are usually multiple orders in a user account, and it is necessary for operations on orders such as retrieval, read or cancel. However, unique identifiers are unnecessary when the data source is singular, eliminating the need for distinction. For example, for an online shopping toolkit, a `cart\_id` for a user's shopping cart or a `profile\_id` for the user's own profile are redundant because a user account usually contains only one of each.\\6. Stored data access requirements: For cases where some tools in the toolkit require specific user details such as address or payment method, and it is reasonable to assume that the user has previously provided this information to the toolkit, there must be a tool API that can access the stored data. For example, in a toolkit for food delivery or e-commerce, there should be a tool API for accessing a user's saved addresses and payment methods.\\7. Operation Status Indication Requirement: For tool APIs that involve operations such as creating, updating, or deleting data resources, it is critical to include the operation status (e.g., a Boolean value of "success") as part of the return value. Other return values should also align with the operation status. For example, if a tool returns a unique identifier for a created data resource, it should return null or an empty string if the operation failed.\\8. File/Media Management Requirement: Toolkits that involve file or media resources (such as the Twitter toolkit that requires media attachments) need to carefully distinguish between local and remote file management. Remote files located on a remote server should be identified and managed by their remote file path or unique identifier. Whereas local files located on the user's system should be identified and managed by local file paths. Different tools may require remote or local files as input parameters or return values, depending on the specific use case. For example, for the Twitter toolkit, a tool for posting a tweet should require the local path of the media file to be posted, while a tool for retrieving tweets should return the remote file identifier of the media file in the tweet. Additionally, there should be a tool for downloading remote media files to the local system.

\noindent \#\#\# Output Format\\Toolkit specifications should be in a valid JSON list separated by ```json` and ```. Each item in the list should contain the following fields:\\* `tool`: The name of the tool, which should be in "CamelCase" format.\\* `name`: The Chinese name of the tool, which should express the function of the tool as concisely as possible.\\* `description`: The summary of the tool, which should be a clear and concise description of the purpose and function of the tool without any ambiguity. It must be in Chinese.\\* `parameters`: The parameter list of the tool, each parameter should contain the fields including `name`, `type` and `description`, `required` (whether the parameter is required). `description` must be in Chinese.\\* `returns`: The return list of the tool, each return should contain `name`, `type` and `description`.\\\\Please note:\\1. For tool `parameters` and `returns`, `name` should not contain spaces and should be in "snake\_case" format.\\2. For tool `parameters` and `returns`, `type` should be a valid JSON type, i.e. one of \texttt{["string", "integer", "number", "boolean", "array", "object", "null"]}.\\3. For tool `parameters` and `returns`, `description` should be a clear and concise description of the parameter or return, without any ambiguity. If the parameter or return is subject to some specific format or value constraints, these constraints should be explicitly specified in `description`.\\4. If a tool `parameters` or `returns` is an object, the exact fields of the object should be explicitly specified in `description`.

\noindent \#\#\# Format example\\You should output the toolkit as a JSON object, strictly following the structure shown in the following example tool spec:\\**WARNING! ** You need to check your JSON format carefully!!! Output the detailed toolkit spec as:\\```json\\<Output toolkit spec follows \texttt{[format description]}>\\```\\Try to generate tools that users might use in their daily lives.\\Do not output the same string repeatedly!!!\\Make sure to meet the above format requirements, especially do not lose "parameters"\\
\noindent Question: You need to generate tools for the following toolkits related to map navigation: Reference message: Turn on navigation sound, turn off navigation sound, change navigation sound, query real-time bus, view my footprints, find charging piles, add common addresses, view common addresses, delete common addresses, add favorite places, view favorite places, delete favorite places, add favorite routes, view favorite routes, delete common routes
\end{boiteepaisseavecuntitre}

\subsection{Query Generation}
\label{appx:Query Generation}
\begin{boiteepaisseavecuntitre}{Generating Single-turn Queries}
\noindent You are now a mobile phone user. Your task is to fill in the corresponding parameters according to the actual scenario based on the provided description and parameters, and then generate a user request based on the filled parameters.\\\\Please make sure that the generated user requests are different from each other. You can feel free to use different sentence styles, such as instructions or requests, and change the degree of detail as needed.

\noindent Related requirements:\\1. Use as few questions and polite tones as possible;\\2. Natural questions similar to human conversations need to be more humane and colloquial, and in line with people's daily life\\3. The parameters in the generated user request should be as reasonable as possible. For example, if it involves image parameters, the user request needs to have representative information like test.jpg\\4. Fill in parameters carefully, output the corresponding JSON format and wrap it with ```json...```\\5. Information such as test.jpg contained in parameters also needs to be included in the user request!!\\6. All parameters must be filled in all given parameters!! Do not miss any one!\\7. The user request needs to be wrapped with <user0>...<user1>. Your task is to generate natural and colloquial user requests, make sure to clearly mention the tool name \{name\_for\_human\} in the request, and avoid using stiff or direct imperative tones. For example, do not generate requests like 'Dianping, show me my group purchases', but use more humane and daily conversation expressions, such as 'Help me see the group purchases on Dianping'. Please generate a user request that meets the specifications according to this requirement.\\8. You can generate parameters first, and then generate user requests based on the generated parameters.\\9. The text descriptions in parameters must be in Chinese!\\10. For time parameters, please use Chinese text! And the expressions before and after must be consistent!\\11. When no parameter definition is given, please generate a statement that does not contain any parameters!
\noindent Output Format:\\1. Parameters: ```json parameter dictionary```. <user0>\texttt{[user request]}<user1>\\2. Parameters: ```json parameter dictionary```. <user0>\texttt{[user request]}<user1>\\…
\noindent Here are two examples for your reference:\\description: Modify alarm time, ring mode, weekday settings, custom settings\\parameters:\[\]\\Must include parameters:\\Output:\\1. Parameters:```json\{\}```. <user0>I want to modify the alarm time<user1>\\\\description: Modify the alarm time, ring mode, weekday settings, custom settings\\parameters:\texttt{[\{"name": "clock\_time","type": "string","description": "Alarm time","required": false\},\{"name": "target\_time","type": "string","description": "Modified target time","required": true\}]}\\Must include parameters: clock\_time, target\_time
Output:\\1. Parameters:```json\{"clock\_time": "8 o'clock","target\_time": "15 o'clock"\}```. <user0>Change the alarm at 8 o'clock to 15 o'clock in the afternoon<user1>
\noindent The description of the tool and the corresponding parameters are as follows:\\description:\{description\}\\parameters:\{parameters\}\\Must include parameters:\{param\_list\}
\noindent Please refer to the example and generate \{num\} \texttt{[user requests]} and \texttt{[parameter dictionaries]} that conform to the output format according to the above requirements and tool conditions.
\end{boiteepaisseavecuntitre}

\subsection{Parameters Selection}
\label{appx:Parameters Selection}
\begin{boiteepaisseavecuntitre}{Selecting parameters}
\noindent Please select the specified number of parameters that best suit your daily habits based on the tool description and corresponding parameter definitions.
\noindent \#\# Selection Principles:\\1. Simulate daily user usage scenarios and give priority to parameters that are more commonly used in daily life
\noindent \#\# Output Format:\\1. Output your thoughts first, then output the parameter name you selected, and wrap it with ```...```\\2. Use commas to separate parameter names
\noindent \#\# For examples:\\Tool description: Create a new alarm\\Parameter definition:\\\texttt{[{"name": "clock\_time\_name","type": "string","description": "Alarm name, empty by default","required": false},{"name": "clock\_time","type": "string","description": "Alarm time","required": true},{"name": "kwargs","type": "array","description": "Ring mode, weekday settings, custom settings","required": false}]}\\Output 2 parameters: clock\_time, clock\_time\_name\\\\Start now!
\noindent Tool description: \{tool\_desc\}\\Parameter definition:
\{parameters\}\\Output \{num\} parameters:
\end{boiteepaisseavecuntitre}

\subsection{LLM Check Prompt}
\label{appx:LLM Check Prompt}
\begin{boiteepaisseavecuntitre}{LLM Check Prompt}
\noindent \#\# System
\\You are an expert who can accurately judge the information matching relationship between texts, and you have a keen insight into the information matching between two structured texts.

\noindent \#\# Task Description
\\1. Given a query, model\_output and the label of the corresponding tool call, your task is to determine whether the model\_output and label both express the same semantic meaning and are derived from the information found in the query. output your judgment True/False\\
\\2. If model\_output and label have different number of parameters, such as \{'A':1,'B':2\} and \{'A':1\}, output False
\\3. model\_output and label do not have to be exactly the same. As long as they express the same meaning and can solve query needs, only True is output.
\\4. If model\_output is the same as label, or only a slight difference between them in terms of singular and plural, output True! such as \{\} and \{\}, \{"departure": "office"\} and \{"departure": "Offices"\}, you need output True

\noindent \#\# Output Format
\\1. Following Task Description, Only output True/False, Never answer more text!
\noindent For examples:
\\1. Different characters but same semantics, output True:
\\query: Add a note of the address of Starbucks Zhongguancun store on the map, and write "frequently visited writing place"
\\label: \{"address": "Starbucks Zhongguancun store", "remake": "often come to write"\}
\\model\_output: \{"address": "Starbucks Zhongguancun store", "remake": "frequently visited writing place"\}
\\Output: True
\\2. model\_output contains parameters that are not in label, output False:
\\query: How many times have you ridden this year?
\\label: \{\}
\\model\_output: \{"time": "this year"\}
\\answer: False
\\3. The model\_output is missing the parameters of label, output False:
\\query: I want to know which five-star hotels are there in Nanshan District, Shenzhen
\\label: \{"destination": "Nanshan District, Shenzhen", "checkin\_date":"", "checkout\_date":"", "price\_range":"", "kwargs":"five-star hotel"\}
\\model\_output: \{"destination": "Nanshan District, Shenzhen", "checkin\_date":"", "checkout\_date":"", "kwargs":"five-star hotel"\}
\\answer: False
\\4. Time hallucination occurs, output False:
\\query: Make an appointment to test drive Xiaomi su7 at 10 am this Saturday
\\label: \{"test\_drive\_time": "This Saturday at 10 am", "name":"", "phone":"", "kwargs":""\}
\\model\_output: \{"test\_drive\_time": "2023-04-20 10:00", "name":"", "phone":"", "kwargs":""\}
answer: False
\\5. The time format is different, but it expresses the same time, output True:
\\query: Make an appointment to test drive Xiaomi su7 at 10 am this Saturday
\\label: \{"test\_drive\_time": "This Saturday at 10 am", "name":"", "phone":"", "kwargs":""\}
\\model\_output: \{"test\_drive\_time": "This Saturday at 10:00", "name":"", "phone":"", "kwargs":""\}
\\answer: True
\\6. All are empty characters, output True:
\\query: Open TikTok
\\label: \{\}
\\model\_output: \{\}
\\answer: True

\noindent Start now!\\\\query: \{query\}\\label: \{label\}\\model\_output: \{model\_output\}\\answer:
\end{boiteepaisseavecuntitre}

\subsection{Irrelevant tools selection}
\label{appx:Irrelevant prompt}
\begin{boiteepaisseavecuntitre}{Irrelevant tools selection}
\noindent Given a query and the corresponding tool description and parameter definition, please judge whether the tool can complete the query.\\
\noindent \#\# Output Format:\\Only outputs boolean strings: True/False\\\\Start now!\\query:\{query\}\\tool:\{tool\}\\answer:
\end{boiteepaisseavecuntitre}

\subsection{Generation with external individual information}
\label{appx:external individual information Prompt}
\begin{boiteepaisseavecuntitre}{classify entity}
\noindent Please determine whether a parameter is in the entity category\\
\noindent \#\# Output Format:\\Please select which entity it belongs to. The selected entity category is enclosed in ```...```\\entity:\{entities\}\\Parameter name and description:\{param\_desc\}\\Output:
\end{boiteepaisseavecuntitre}
\begin{boiteepaisseavecuntitre}{pronoun generation}
\noindent Please rewrite the following content into a sentence in the form of adjective + noun that is more in line with daily habits and has more specific information.\\\\For example:\\Input: The time mentioned in the message received from SMS\\Requirement: Cannot contain specific time\\Output: The time mentioned in the text message sent to me by Tom\\\\Start now! Output only the rewritten 1 sentence!\\\\Input: \{entity\} mentioned in the message received from \{Tool\_\}
Requirement: Cannot contain specific \{entity\}
Output:
\end{boiteepaisseavecuntitre}
\begin{boiteepaisseavecuntitre}{external individual information simulation}
\noindent Please simulate a message in \{app\} as required\\\\
\noindent \#\# Related requirements\\The content must involve {entity};\\The content should be as detailed as possible, and no referential content such as that or this should appear. The message should contain a clear \{entity\}: \{param\_value\}
The content must not contain the following: \{param\_forbid\}
\noindent \#\# Output format\\Only output simulated messages, do not ask or output other content!\\\\Simulated message:
\end{boiteepaisseavecuntitre}
\begin{boiteepaisseavecuntitre}{rewrite query}
\noindent Please generate the corresponding user query according to the label and the corresponding tool function description

\noindent \#\# Related requirements\\1. Try to be consistent with the language of daily conversation interaction, smooth and concise

\noindent \#\# Output format:\\Only output the query, do not include other information\\\\Tool description: \{desc\}\\label: \{label\}\\query:
\end{boiteepaisseavecuntitre}

\subsection{Multi-turn Prompt}
\label{appx:Multi-turn Prompt}
\begin{boiteepaisseavecuntitre}{Question prompt}
\noindent You are a function call agent, currently processing the \{description\} function call task, but the parameters \{miss\_args\_desc\} are missing. Please ask the user for the missing parameter values\\\\

\noindent **Requirements**\\
- Do not include the existing parameters: \{args\_desc\}\\
- Be brief and concise, do not add additional explanations\\
- Ask only about the missing parameters\\
- Ask multiple missing parameters together, do not use special symbols to separate, such as colons, newlines, etc.\\
- Just ask questions, do not tell the user how to answer
\end{boiteepaisseavecuntitre}

\newpage

\section{Extra Experiment Details}
\label{appx: more res}

Our experiments are all conducted in the \hb datasets generated by open-source LLMs. When generating these data, we only summarized the behaviors and scenarios based on anonymous user logs, without involving any privacy information of users. And \hb's evaluations strictly follow the license of the corresponding LLMs. Qwen2.5-72B-Instruct, Qwen2.5-7B-Instruct(\citealp{qwen2.5}) and ToolACE-8B(\citealp{liu2024toolace}) are Apache-2.0 LICENSE; xLAM-7b-fc-r(\citealp{liuapigen}) is CC-by-NC-4.0 LICENSE; Llama-3.1-70B-Instruct(\citealp{dubey2024llama}) and Llama-3.1-8B-Instruct are llama3.1 LICENSE; Ministral-8B-Instruct(\citealp{mistral}) is mrl LICENSE. They all allow developers to use their technology for non-commercial purposes and to support research work. And GPT-4o(\citealp{hurst2024gpt}) and Claude3.5-sonnet(\citealp{anthropic2024introducing}) are also widely used in research. Without using LLM judge, it takes approximately 90 minutes for the 7B LLMs to inference all single turn(6500+ samples) and multi turn data(6400+ samples) using a single A1OO 80G GPU. As for GPT-4o, one evaluation approximately cost 75 dollars.

The experiment result of single-turn dataset is shown in \Cref{table:single turn}. As for single-turn dataset with missing arguments (Imperfect), PHR is much larger than other single-turn dataset. It reveals that user query with missing arguments can easily lead to parameter hallucinations. In these cases, LLMs tend to fill in missing arguments based on their internal model of the world rather than the actual user input, as detailed in \Cref{appx:badcase param hallucination}. Since snapshots are often imperfect, these inaccuracies significantly impact multi-turn success rates, further highlighting the challenges posed by incomplete or unclear user input. As a result, the overall effectiveness of function-calling tasks is diminished when the instructions provided to the model are less than ideal. And the single-turn dataset with external individual information has the lower end-to-end \textbf{Acc.} .

We also evaluate function-calling snapshots across three types for detailed investigation. (1) every turn throughout the conversation in \Cref{table:MT all snapshot}, (2) the turn when the user changes slot values in \Cref{table:MT SO EK}, and (3) the first and last turn of the conversation in \Cref{table:MT first last}. Some additional observations are as follows.

\textbf{Fluctuations in Conversation Success Rate Across Diverse Q\&A Trajectories:} As shown in \Cref{table:MT all snapshot}, different conversation trajectories introduce varying degrees of disturbance to the overall success rate of function calling. Consistent with expectations, we find that the progress rate tends to be higher when slot values are provided more rapidly within the conversation (e.g. mQmA and sQmA, where users provide multiple slot values at once and thus answer models' questions faster.), highlighting the impact of the timing and flow of user input on function-calling accuracy.

\textbf{Generalization and Handling of Argument Shifts and External information:} As shown in \Cref{table:MT SO EK}, open-source LLMs are generally slow to detect and adjust to slot overriding. Only GPT-4o and Claude3.5-sonnet, with its more robust generalization abilities, is able to perform this task more effectively. As for external individual information, external noise increases Parameter Mismatch Rate (PMR) and causes the model to miss crucial arguments, hindering its ability to fill slots correctly. 

\textbf{Imperfect Instructions and Parameter Name Hallucinations:} As illustrated in \Cref{table:MT first last}, the PHR in the first snapshot is significantly higher than in the last snapshot for all LLMs, with the PHR being positively correlated with the rate of parameter name hallucinations. Furthermore, when all parameter names are predicted correctly, the end-to-end function call is typically accurate as well, i.e. Other Error Rate$\approx 0\%$, emphasizing the importance of precise initial instruction. We can see similar observations from \Cref{table:single turn}.

\begin{table*}[htb]
\centering
\scriptsize
\setlength\tabcolsep{4pt}
\begin{tabular}{ccccccc}
\toprule
Data type                       & Model                               & Func. Acc. & PHR  & PMR  & Acc. & Irrelevant \\ \midrule
\multirow{10}{*}{Perfect}   & GPT-4o (Prompt)                     & 89.74\%  & 0.16\%  & 10.63\% & 78.69\%  & 64.38\%    \\
                                & Claude3.5-sonnet (Prompt)           & 89.89\%  & 0.26\%  & 7.72\%  & 79.86\%  & 69.35\%    \\
                                & Qwen2.5-72B-Instruct (Prompt)       & 90.50\%  & 1.98\%  & 3.92\%  & 80.86\%  & 74.59\%    \\
                                & Llama-3.1-70B-Instruct (Prompt)     & 91.87\%  & 6.64\%  & 4.27\%  & 75.99\%  & 14.19\%    \\
                                & Qwen2.5-7B-Instruct (Prompt)        & 89.41\%  & 5.50\%  & 3.96\%  & 75.75\%  & 37.36\%    \\
                                & Llama-3.1-8B-Instruct (Prompt)      & 87.75\%  & 5.22\%  & 3.39\%  & 73.01\%  & 9.93\%     \\
                                & Ministral-8B-Instruct-2410 (Prompt) & 88.32\%  & 0.43\%  & 4.23\%  & 78.02\%  & 1.14\%     \\
                                & Hammer2.1-7b (FC)                   & 80.29\%  & 1.06\%  & 7.95\%  & 68.85\%  & 89.35\%    \\
                                & xLAM-7b-fc-r (FC)                   & 83.17\%  & 0.34\%  & 5.00\%  & 73.48\%  & 49.76\%    \\
                                & ToolACE-8B (FC)                     & 91.44\%  & 0.78\%  & 4.96\%  & 78.82\%  & 22.98\%    \\ \midrule
\multirow{10}{*}{Imperfect} & GPT-4o (Prompt)                     & 88.86\%  & 7.61\%  & 4.48\%  & 78.77\%  & 71.17\%    \\
                                & Claude3.5-sonnet (Prompt)           & 88.94\%  & 8.15\%  & 2.49\%  & 79.38\%  & 67.90\%    \\
                                & Qwen2.5-72B-Instruct (Prompt)       & 84.59\%  & 16.27\% & 1.68\%  & 68.64\%  & 78.97\%    \\
                                & Llama-3.1-70B-Instruct (Prompt)     & 91.20\%  & 20.87\% & 1.90\%  & 69.47\%  & 19.04\%    \\
                                & Qwen2.5-7B-Instruct (Prompt)        & 82.09\%  & 26.72\% & 2.52\%  & 58.08\%  & 49.04\%    \\
                                & Llama-3.1-8B-Instruct (Prompt)      & 89.35\%  & 40.72\% & 1.80\%  & 51.08\%  & 8.50\%     \\
                                & Ministral-8B-Instruct-2410 (Prompt) & 75.03\%  & 54.17\% & 2.59\%  & 32.19\%  & 6.09\%     \\
                                & Hammer2.1-7b (FC)                   & 81.88\%  & 24.38\% & 3.02\%  & 59.29\%  & 89.23\%    \\
                                & xLAM-7b-fc-r (FC)                   & 86.45\%  & 41.05\% & 2.61\%  & 48.82\%  & 54.51\%    \\
                                & ToolACE-8B (FC)                     & 82.53\%  & 33.58\% & 3.96\%  & 51.94\%  & 29.37\%    \\ \midrule
\multirow{10}{*}{External}  & GPT-4o (Prompt)                     & 87.91\%  & 2.81\%  & 29.11\% & 56.16\%  & 73.02\%    \\
                                & Claude3.5-sonnet (Prompt)           & 82.13\%  & 2.49\%  & 25.27\% & 55.83\%  & 67.23\%    \\
                                & Qwen2.5-72B-Instruct (Prompt)       & 93.10\%  & 5.58\%  & 12.52\% & 67.40\%  & 65.70\%    \\
                                & Llama-3.1-70B-Instruct (Prompt)     & 93.70\%  & 5.72\%  & 14.35\% & 59.57\%  & 4.26\%     \\
                                & Qwen2.5-7B-Instruct (Prompt)        & 88.51\%  & 9.90\%  & 8.94\%  & 58.89\%  & 36.59\%    \\
                                & Llama-3.1-8B-Instruct (Prompt)      & 89.87\%  & 9.94\%  & 8.52\%  & 57.61\%  & 5.96\%     \\
                                & Ministral-8B-Instruct-2410 (Prompt) & 90.12\%  & 11.89\% & 6.70\%  & 57.70\%  & 4.34\%     \\
                                & Hammer2.1-7b (FC)                   & 77.87\%  & 7.65\%  & 13.44\% & 50.21\%  & 90.63\%    \\
                                & xLAM-7b-fc-r (FC)                   & 86.89\%  & 10.08\% & 15.08\% & 48.85\%  & 45.78\%    \\
                                & ToolACE-8B (FC)                     & 92.25\%  & 11.71\% & 10.70\% & 56.93\%  & 27.40\%    \\ \midrule
\multirow{10}{*}{Overall}       & GPT-4o (Prompt)                     & 88.83\%  & 3.52\%  & 14.74\% & 71.21\%  & 69.52\%    \\
                                & Claude3.5-sonnet (Prompt)           & 86.99\%  & 3.63\%  & 11.82\% & 71.69\%  & 68.16\%    \\
                                & Qwen2.5-72B-Instruct (Prompt)       & 89.40\%  & 7.94\%  & 6.04\%  & 72.30\%  & 73.09\%    \\
                                & Llama-3.1-70B-Instruct (Prompt)     & 92.25\%  & 11.07\% & 6.84\%  & 68.34\%  & 12.49\%    \\
                                & Qwen2.5-7B-Instruct (Prompt)        & 86.67\%  & 14.04\% & 5.14\%  & 64.24\%  & 41.00\%    \\
                                & Llama-3.1-8B-Instruct (Prompt)      & 88.99\%  & 18.63\% & 4.57\%  & 60.57\%  & 8.13\%     \\
                                & Ministral-8B-Instruct-2410 (Prompt) & 84.49\%  & 22.16\% & 4.51\%  & 55.97\%  & 3.86\%     \\
                                & Hammer2.1-7b (FC)                   & 80.01\%  & 11.03\% & 8.13\%  & 59.45\%  & 89.74\%    \\
                                & xLAM-7b-fc-r (FC)                   & 85.50\%  & 17.16\% & 7.56\%  & 57.05\%  & 50.02\%    \\
                                & ToolACE-8B (FC)                     & 88.74\%  & 15.35\% & 6.54\%  & 62.56\%  & 26.58\%    \\ \bottomrule
\end{tabular}
\caption{Experiment result for single-turn dataset.}
\setlength{\abovecaptionskip}{0cm}
\setlength{\belowcaptionskip}{0cm}
\label{table:single turn}
\end{table*}

\begin{table*}[htb]
\centering
\tiny
\begin{tabular}{cccccccc}
\toprule
Data type                      & Model                               & Func. Acc. & PHR  & PMR  & Acc. & PR      & SR      \\ \midrule
\multirow{10}{*}{sQsA}         & GPT-4o (Prompt)                     & 89.85\%  & 9.31\%  & 5.11\%  & 72.52\%  & 66.03\% & 54.95\% \\
                               & Claude3.5-sonnet (Prompt)           & 91.08\%  & 9.02\%  & 4.28\%  & 76.62\%  & 70.64\% & 62.37\% \\
                               & Qwen2.5-72B-Instruct (Prompt)       & 82.17\%  & 12.29\% & 6.35\%  & 64.59\%  & 56.28\% & 44.55\% \\
                               & Llama-3.1-70B-Instruct (Prompt)     & 91.83\%  & 12.86\% & 1.71\%  & 75.05\%  & 64.72\% & 55.19\% \\
                               & Qwen2.5-7B-Instruct (Prompt)        & 82.67\%  & 19.50\% & 5.23\%  & 60.08\%  & 48.90\% & 37.62\% \\
                               & Llama-3.1-8B-Instruct (Prompt)      & 90.59\%  & 22.87\% & 19.18\% & 51.53\%  & 38.43\% & 18.56\% \\
                               & Ministral-8B-Instruct-2410 (Prompt) & 71.28\%  & 33.17\% & 4.81\%  & 43.19\%  & 26.78\% & 20.29\% \\
                               & Hammer2.1-7b (FC)                   & 81.68\%  & 15.71\% & 6.13\%  & 62.26\%  & 52.37\% & 41.08\% \\
                               & xLAM-7b-fc-r (FC)                   & 91.33\%  & 29.19\% & 6.81\%  & 58.03\%  & 39.28\% & 32.17\% \\
                               & ToolACE-8B (FC)                     & 80.69\%  & 28.11\% & 3.11\%  & 54.54\%  & 42.55\% & 26.98\% \\ \midrule
\multirow{10}{*}{mQmA}         & GPT-4o (Prompt)                     & 90.84\%  & 7.12\%  & 5.60\%  & 75.63\%  & 73.22\% & 64.35\% \\
                               & Claude3.5-sonnet (Prompt)           & 90.84\%  & 7.93\%  & 4.43\%  & 75.52\%  & 73.74\% & 65.59\% \\
                               & Qwen2.5-72B-Instruct (Prompt)       & 81.93\%  & 10.58\% & 2.71\%  & 67.82\%  & 64.75\% & 57.67\% \\
                               & Llama-3.1-70B-Instruct (Prompt)     & 91.58\%  & 12.22\% & 2.54\%  & 73.20\%  & 67.42\% & 59.40\% \\
                               & Qwen2.5-7B-Instruct (Prompt)        & 82.42\%  & 20.64\% & 5.29\%  & 58.12\%  & 53.06\% & 44.55\% \\
                               & Llama-3.1-8B-Instruct (Prompt)      & 90.59\%  & 23.65\% & 10.83\% & 57.70\%  & 51.15\% & 40.59\% \\
                               & Ministral-8B-Instruct-2410 (Prompt) & 71.28\%  & 35.34\% & 6.48\%  & 40.29\%  & 28.59\% & 21.28\% \\
                               & Hammer2.1-7b (FC)                   & 82.17\%  & 14.87\% & 8.67\%  & 60.86\%  & 57.57\% & 46.53\% \\
                               & xLAM-7b-fc-r (FC)                   & 91.33\%  & 33.79\% & 7.64\%  & 52.53\%  & 40.74\% & 34.40\% \\
                               & ToolACE-8B (FC)                     & 80.44\%  & 27.54\% & 4.96\%  & 52.63\%  & 46.39\% & 33.41\% \\ \midrule
\multirow{10}{*}{mQsA}         & GPT-4o (Prompt)                     & 90.34\%  & 9.17\%  & 7.88\%  & 73.00\%  & 68.52\% & 59.15\% \\
                               & Claude3.5-sonnet (Prompt)           & 90.59\%  & 9.15\%  & 6.42\%  & 73.34\%  & 68.93\% & 59.65\% \\
                               & Qwen2.5-72B-Instruct (Prompt)       & 81.93\%  & 13.74\% & 5.11\%  & 63.90\%  & 56.93\% & 47.27\% \\
                               & Llama-3.1-70B-Instruct (Prompt)     & 91.33\%  & 15.07\% & 3.68\%  & 70.74\%  & 61.63\% & 52.47\% \\
                               & Qwen2.5-7B-Instruct (Prompt)        & 82.17\%  & 25.62\% & 6.76\%  & 54.40\%  & 43.62\% & 32.42\% \\
                               & Llama-3.1-8B-Instruct (Prompt)      & 90.59\%  & 27.01\% & 19.78\% & 48.59\%  & 38.42\% & 19.05\% \\
                               & Ministral-8B-Instruct-2410 (Prompt) & 71.28\%  & 45.38\% & 7.79\%  & 32.94\%  & 19.02\% & 8.91\%  \\
                               & Hammer2.1-7b (FC)                   & 81.68\%  & 26.24\% & 11.26\% & 52.63\%  & 41.50\% & 25.24\% \\
                               & xLAM-7b-fc-r (FC)                   & 91.33\%  & 40.49\% & 10.62\% & 46.00\%  & 30.36\% & 19.80\% \\
                               & ToolACE-8B (FC)                     & 80.44\%  & 47.84\% & 7.76\%  & 38.55\%  & 25.87\% & 8.42\%  \\ \midrule
\multirow{10}{*}{sQmA}         & GPT-4o (Prompt)                     & 91.83\%  & 8.22\%  & 7.18\%  & 74.05\%  & 70.79\% & 59.90\% \\
                               & Claude3.5-sonnet (Prompt)           & 91.33\%  & 7.77\%  & 6.84\%  & 73.94\%  & 72.62\% & 62.12\% \\
                               & Qwen2.5-72B-Instruct (Prompt)       & 82.42\%  & 11.42\% & 2.82\%  & 67.40\%  & 64.19\% & 56.68\% \\
                               & Llama-3.1-70B-Instruct (Prompt)     & 91.83\%  & 11.85\% & 3.68\%  & 73.52\%  & 68.00\% & 60.39\% \\
                               & Qwen2.5-7B-Instruct (Prompt)        & 82.17\%  & 20.46\% & 5.18\%  & 57.70\%  & 53.04\% & 44.55\% \\
                               & Llama-3.1-8B-Instruct (Prompt)      & 90.59\%  & 24.35\% & 11.18\% & 56.22\%  & 50.35\% & 39.10\% \\
                               & Ministral-8B-Instruct-2410 (Prompt) & 71.28\%  & 36.08\% & 6.19\%  & 40.18\%  & 28.37\% & 21.03\% \\
                               & Hammer2.1-7b (FC)                   & 81.93\%  & 14.13\% & 8.95\%  & 60.65\%  & 57.18\% & 46.28\% \\
                               & xLAM-7b-fc-r (FC)                   & 91.33\%  & 33.79\% & 7.75\%  & 52.63\%  & 41.08\% & 34.65\% \\
                               & ToolACE-8B (FC)                     & 80.69\%  & 27.60\% & 6.51\%  & 51.68\%  & 46.10\% & 32.42\% \\ \midrule
\multirow{10}{*}{w/ SO}        & GPT-4o (Prompt)                     & 90.37\%  & 8.52\%  & 4.90\%  & 72.98\%  & 68.65\% & 57.93\% \\
                               & Claude3.5-sonnet (Prompt)           & 90.44\%  & 8.83\%  & 3.58\%  & 76.49\%  & 72.54\% & 65.01\% \\
                               & Qwen2.5-72B-Instruct (Prompt)       & 85.08\%  & 13.62\% & 5.08\%  & 65.40\%  & 60.25\% & 50.17\% \\
                               & Llama-3.1-70B-Instruct (Prompt)     & 91.47\%  & 16.17\% & 1.56\%  & 70.25\%  & 62.01\% & 52.09\% \\
                               & Qwen2.5-7B-Instruct (Prompt)        & 82.88\%  & 23.21\% & 3.92\%  & 54.21\%  & 45.78\% & 33.05\% \\
                               & Llama-3.1-8B-Instruct (Prompt)      & 90.10\%  & 27.30\% & 14.56\% & 48.84\%  & 37.99\% & 22.54\% \\
                               & Ministral-8B-Instruct-2410 (Prompt) & 74.22\%  & 37.75\% & 3.60\%  & 40.29\%  & 26.21\% & 19.93\% \\
                               & Hammer2.1-7b (FC)                   & 82.95\%  & 19.92\% & 5.03\%  & 57.85\%  & 49.83\% & 36.83\% \\
                               & xLAM-7b-fc-r (FC)                   & 89.89\%  & 34.70\% & 6.51\%  & 48.37\%  & 34.44\% & 24.94\% \\
                               & ToolACE-8B (FC)                     & 81.51\%  & 32.96\% & 3.05\%  & 50.25\%  & 40.07\% & 26.87\% \\ \midrule
\multirow{10}{*}{w/ mSv}       & GPT-4o (Prompt)                     & 90.68\%  & 7.84\%  & 6.45\%  & 61.98\%  & 56.45\% & 18.53\% \\
                               & Claude3.5-sonnet (Prompt)           & 90.02\%  & 8.02\%  & 5.36\%  & 69.68\%  & 63.94\% & 42.89\% \\
                               & Qwen2.5-72B-Instruct (Prompt)       & 86.17\%  & 12.23\% & 5.66\%  & 64.98\%  & 57.35\% & 42.05\% \\
                               & Llama-3.1-70B-Instruct (Prompt)     & 91.62\%  & 14.44\% & 5.21\%  & 68.87\%  & 58.64\% & 41.39\% \\
                               & Qwen2.5-7B-Instruct (Prompt)        & 83.06\%  & 21.07\% & 4.85\%  & 51.85\%  & 41.75\% & 18.15\% \\
                               & Llama-3.1-8B-Instruct (Prompt)      & 89.84\%  & 25.19\% & 13.85\% & 48.30\%  & 34.97\% & 15.80\% \\
                               & Ministral-8B-Instruct-2410 (Prompt) & 75.25\%  & 33.64\% & 4.41\%  & 36.47\%  & 22.00\% & 7.53\%  \\
                               & Hammer2.1-7b (FC)                   & 83.25\%  & 17.78\% & 10.81\% & 52.81\%  & 43.87\% & 17.49\% \\
                               & xLAM-7b-fc-r (FC)                   & 89.18\%  & 26.90\% & 15.51\% & 43.23\%  & 29.09\% & 4.80\%  \\
                               & ToolACE-8B (FC)                     & 81.93\%  & 29.82\% & 3.82\%  & 45.74\%  & 35.07\% & 11.57\% \\ \midrule
\multirow{10}{*}{w/ External}  & GPT-4o (Prompt)                     & 90.81\%  & 5.52\%  & 13.36\% & 69.64\%  & 66.34\% & 49.18\% \\
                               & Claude3.5-sonnet (Prompt)           & 90.40\%  & 6.22\%  & 9.18\%  & 71.83\%  & 68.62\% & 54.28\% \\
                               & Qwen2.5-72B-Instruct (Prompt)       & 84.48\%  & 12.36\% & 8.98\%  & 61.12\%  & 59.58\% & 45.91\% \\
                               & Llama-3.1-70B-Instruct (Prompt)     & 92.04\%  & 12.54\% & 3.77\%  & 69.29\%  & 63.47\% & 49.79\% \\
                               & Qwen2.5-7B-Instruct (Prompt)        & 81.22\%  & 19.73\% & 6.35\%  & 53.87\%  & 50.04\% & 34.89\% \\
                               & Llama-3.1-8B-Instruct (Prompt)      & 89.79\%  & 22.70\% & 14.95\% & 50.14\%  & 43.56\% & 25.51\% \\
                               & Ministral-8B-Instruct-2410 (Prompt) & 73.67\%  & 33.52\% & 9.88\%  & 35.00\%  & 25.06\% & 14.08\% \\
                               & Hammer2.1-7b (FC)                   & 82.85\%  & 15.74\% & 17.72\% & 48.73\%  & 46.32\% & 25.30\% \\
                               & xLAM-7b-fc-r (FC)                   & 88.77\%  & 26.38\% & 13.66\% & 48.30\%  & 39.76\% & 25.91\% \\
                               & ToolACE-8B (FC)                     & 83.87\%  & 28.83\% & 5.45\%  & 49.15\%  & 43.61\% & 26.93\% \\ \midrule
\multirow{10}{*}{Overall} & GPT-4o (Prompt) & 90.67 & 7.96 & 7.21 & 71.40 & 67.14 & 52.00 \\
& Claude3.5-sonnet (Prompt) & 90.67 & 8.13 & 5.73 & 73.92 & 70.15 & 58.84 \\
& Qwen2.5-72B-Instruct (Prompt) & 83.45 & 12.32 & 5.24 & 65.03 & 59.90 & 49.19 \\
& Llama-3.1-70B-Instruct (Prompt) & 91.67 & 13.59 & 3.16 & 71.56 & 63.70 & 52.96 \\
& Qwen2.5-7B-Instruct (Prompt) & 82.37 & 21.46 & 5.37 & 55.75 & 48.03 & 35.03 \\
& Llama-3.1-8B-Instruct (Prompt) & 90.30 & 24.72 & 14.90 & 51.62 & 42.12 & 25.88 \\
& Ministral-8B-Instruct (Prompt) & 72.61 & 36.41 & 6.17 & 38.34 & 25.15 & 16.15 \\
& Hammer2.1-7b (FC) & 82.36 & 17.77 & 9.80 & 56.54 & 49.81 & 34.11 \\
& xLAM-7b-fc-r (FC) & 90.45 & 32.18 & 9.79 & 49.87 & 36.39 & 25.24 \\
& ToolACE-8B (FC) & 81.37 & 31.81 & 4.95 & 48.93 & 39.95 & 23.80 \\
\bottomrule
\end{tabular}
\caption{Overall evaluation on \hb for different metrics of different data types.}
\label{table:Overall}
\end{table*}

\begin{table*}[htb]
\centering
\tiny
\begin{tabular}{cccccccc}
\toprule
Data type                      & Model                               & Func. Acc. & PHR  & PMR  & Acc. & PR      & SR      \\ \midrule
\multirow{10}{*}{w/o SO}       & GPT-4o (Prompt)                     & 89.14\%  & 8.65\%  & 5.68\%  & 72.35\%  & 67.90\% & 58.76\% \\
                               & Claude3.5-sonnet (Prompt)           & 90.51\%  & 8.69\%  & 4.94\%  & 76.03\%  & 71.71\% & 64.39\% \\
                               & Qwen2.5-72B-Instruct (Prompt)       & 84.94\%  & 12.70\% & 5.52\%  & 66.73\%  & 61.11\% & 51.82\% \\
                               & Llama-3.1-70B-Instruct (Prompt)     & 91.75\%  & 14.34\% & 1.97\%  & 73.88\%  & 65.34\% & 58.41\% \\
                               & Qwen2.5-7B-Instruct (Prompt)        & 82.88\%  & 20.27\% & 5.17\%  & 59.71\%  & 50.66\% & 42.06\% \\
                               & Llama-3.1-8B-Instruct (Prompt)      & 90.03\%  & 25.08\% & 15.26\% & 52.66\%  & 41.54\% & 27.69\% \\
                               & Ministral-8B-Instruct-2410 (Prompt) & 74.22\%  & 35.01\% & 5.10\%  & 42.90\%  & 27.76\% & 22.74\% \\
                               & Hammer2.1-7b (FC)                   & 82.68\%  & 16.92\% & 6.58\%  & 61.83\%  & 53.87\% & 44.46\% \\
                               & xLAM-7b-fc-r (FC)                   & 89.69\%  & 30.83\% & 6.55\%  & 55.64\%  & 39.01\% & 33.26\% \\
                               & ToolACE-8B (FC)                     & 81.58\%  & 28.32\% & 3.88\%  & 54.23\%  & 44.08\% & 32.71\% \\ \midrule
\multirow{10}{*}{w/ SO}        & GPT-4o (Prompt)                     & 90.37\%  & 8.52\%  & 4.90\%  & 72.98\%  & 68.65\% & 57.93\% \\
                               & Claude3.5-sonnet (Prompt)           & 90.44\%  & 8.83\%  & 3.58\%  & 76.49\%  & 72.54\% & 65.01\% \\
                               & Qwen2.5-72B-Instruct (Prompt)       & 85.08\%  & 13.62\% & 5.08\%  & 65.40\%  & 60.25\% & 50.17\% \\
                               & Llama-3.1-70B-Instruct (Prompt)     & 91.47\%  & 16.17\% & 1.56\%  & 70.25\%  & 62.01\% & 52.09\% \\
                               & Qwen2.5-7B-Instruct (Prompt)        & 82.88\%  & 23.21\% & 3.92\%  & 54.21\%  & 45.78\% & 33.05\% \\
                               & Llama-3.1-8B-Instruct (Prompt)      & 90.10\%  & 27.30\% & 14.56\% & 48.84\%  & 37.99\% & 22.54\% \\
                               & Ministral-8B-Instruct-2410 (Prompt) & 74.22\%  & 37.75\% & 3.60\%  & 40.29\%  & 26.21\% & 19.93\% \\
                               & Hammer2.1-7b (FC)                   & 82.95\%  & 19.92\% & 5.03\%  & 57.85\%  & 49.83\% & 36.83\% \\
                               & xLAM-7b-fc-r (FC)                   & 89.89\%  & 34.70\% & 6.51\%  & 48.37\%  & 34.44\% & 24.94\% \\
                               & ToolACE-8B (FC)                     & 81.51\%  & 32.96\% & 3.05\%  & 50.25\%  & 40.07\% & 26.87\% \\ \midrule
\multirow{10}{*}{w/o mSv}      & GPT-4o (Prompt)                     & 88.80\%  & 8.37\%  & 5.98\%  & 72.28\%  & 68.47\% & 60.11\% \\
                               & Claude3.5-sonnet (Prompt)           & 90.21\%  & 8.59\%  & 5.32\%  & 75.71\%  & 71.99\% & 65.09\% \\
                               & Qwen2.5-72B-Instruct (Prompt)       & 85.88\%  & 12.73\% & 5.07\%  & 67.83\%  & 63.15\% & 54.84\% \\
                               & Llama-3.1-70B-Instruct (Prompt)     & 91.62\%  & 14.85\% & 2.05\%  & 73.50\%  & 65.70\% & 59.83\% \\
                               & Qwen2.5-7B-Instruct (Prompt)        & 83.16\%  & 20.26\% & 5.23\%  & 60.02\%  & 51.71\% & 44.02\% \\
                               & Llama-3.1-8B-Instruct (Prompt)      & 89.65\%  & 25.98\% & 13.73\% & 52.98\%  & 42.76\% & 31.23\% \\
                               & Ministral-8B-Instruct-2410 (Prompt) & 75.25\%  & 35.55\% & 5.29\%  & 42.89\%  & 28.17\% & 23.61\% \\
                               & Hammer2.1-7b (FC)                   & 83.16\%  & 17.25\% & 6.79\%  & 61.90\%  & 54.76\% & 46.00\% \\
                               & xLAM-7b-fc-r (FC)                   & 89.08\%  & 31.37\% & 6.41\%  & 54.93\%  & 39.14\% & 33.96\% \\
                               & ToolACE-8B (FC)                     & 81.93\%  & 28.45\% & 4.23\%  & 54.24\%  & 44.81\% & 34.90\% \\ \midrule
\multirow{10}{*}{w/ mSv}       & GPT-4o (Prompt)                     & 90.68\%  & 7.84\%  & 6.45\%  & 61.98\%  & 56.45\% & 18.53\% \\
                               & Claude3.5-sonnet (Prompt)           & 90.02\%  & 8.02\%  & 5.36\%  & 69.68\%  & 63.94\% & 42.89\% \\
                               & Qwen2.5-72B-Instruct (Prompt)       & 86.17\%  & 12.23\% & 5.66\%  & 64.98\%  & 57.35\% & 42.05\% \\
                               & Llama-3.1-70B-Instruct (Prompt)     & 91.62\%  & 14.44\% & 5.21\%  & 68.87\%  & 58.64\% & 41.39\% \\
                               & Qwen2.5-7B-Instruct (Prompt)        & 83.06\%  & 21.07\% & 4.85\%  & 51.85\%  & 41.75\% & 18.15\% \\
                               & Llama-3.1-8B-Instruct (Prompt)      & 89.84\%  & 25.19\% & 13.85\% & 48.30\%  & 34.97\% & 15.80\% \\
                               & Ministral-8B-Instruct-2410 (Prompt) & 75.25\%  & 33.64\% & 4.41\%  & 36.47\%  & 22.00\% & 7.53\%  \\
                               & Hammer2.1-7b (FC)                   & 83.25\%  & 17.78\% & 10.81\% & 52.81\%  & 43.87\% & 17.49\% \\
                               & xLAM-7b-fc-r (FC)                   & 89.18\%  & 26.90\% & 15.51\% & 43.23\%  & 29.09\% & 4.80\%  \\
                               & ToolACE-8B (FC)                     & 81.93\%  & 29.82\% & 3.82\%  & 45.74\%  & 35.07\% & 11.57\% \\ \midrule
\multirow{10}{*}{w/o External} & GPT-4o (Prompt)                     & 89.38\%  & 5.22\%  & 3.64\%  & 77.95\%  & 74.36\% & 67.34\% \\
                               & Claude3.5-sonnet (Prompt)           & 90.20\%  & 4.44\%  & 4.36\%  & 80.35\%  & 76.60\% & 70.40\% \\
                               & Qwen2.5-72B-Instruct (Prompt)       & 85.10\%  & 8.98\%  & 5.29\%  & 69.92\%  & 65.94\% & 57.55\% \\
                               & Llama-3.1-70B-Instruct (Prompt)     & 92.44\%  & 9.76\%  & 1.30\%  & 79.57\%  & 72.55\% & 66.93\% \\
                               & Qwen2.5-7B-Instruct (Prompt)        & 81.42\%  & 13.10\% & 3.91\%  & 64.92\%  & 58.13\% & 49.99\% \\
                               & Llama-3.1-8B-Instruct (Prompt)      & 89.79\%  & 17.30\% & 16.13\% & 58.52\%  & 47.80\% & 32.65\% \\
                               & Ministral-8B-Instruct-2410 (Prompt) & 73.67\%  & 28.87\% & 3.39\%  & 47.60\%  & 31.34\% & 26.73\% \\
                               & Hammer2.1-7b (FC)                   & 82.65\%  & 12.33\% & 4.06\%  & 66.19\%  & 60.62\% & 52.85\% \\
                               & xLAM-7b-fc-r (FC)                   & 88.77\%  & 22.59\% & 5.69\%  & 63.09\%  & 49.07\% & 43.46\% \\
                               & ToolACE-8B (FC)                     & 83.87\%  & 23.38\% & 2.68\%  & 60.84\%  & 51.12\% & 39.99\% \\ \midrule
\multirow{10}{*}{w/ External}  & GPT-4o (Prompt)                     & 90.81\%  & 5.52\%  & 13.36\% & 69.64\%  & 66.34\% & 49.18\% \\
                               & Claude3.5-sonnet (Prompt)           & 90.40\%  & 6.22\%  & 9.18\%  & 71.83\%  & 68.62\% & 54.28\% \\
                               & Qwen2.5-72B-Instruct (Prompt)       & 84.48\%  & 12.36\% & 8.98\%  & 61.12\%  & 59.58\% & 45.91\% \\
                               & Llama-3.1-70B-Instruct (Prompt)     & 92.04\%  & 12.54\% & 3.77\%  & 69.29\%  & 63.47\% & 49.79\% \\
                               & Qwen2.5-7B-Instruct (Prompt)        & 81.22\%  & 19.73\% & 6.35\%  & 53.87\%  & 50.04\% & 34.89\% \\
                               & Llama-3.1-8B-Instruct (Prompt)      & 89.79\%  & 22.70\% & 14.95\% & 50.14\%  & 43.56\% & 25.51\% \\
                               & Ministral-8B-Instruct-2410 (Prompt) & 73.67\%  & 33.52\% & 9.88\%  & 35.00\%  & 25.06\% & 14.08\% \\
                               & Hammer2.1-7b (FC)                   & 82.85\%  & 15.74\% & 17.72\% & 48.73\%  & 46.32\% & 25.30\% \\
                               & xLAM-7b-fc-r (FC)                   & 88.77\%  & 26.38\% & 13.66\% & 48.30\%  & 39.76\% & 25.91\% \\
                               & ToolACE-8B (FC)                     & 83.87\%  & 28.83\% & 5.45\%  & 49.15\%  & 43.61\% & 26.93\% \\ \midrule
\end{tabular}
\caption{Ablation on \hb.}
\label{table:Ablation}
\end{table*}

\clearpage

\begin{table*}[ht]
\centering
\small
\setlength\tabcolsep{4pt}
\begin{tabular}{ccccccccccc}
\toprule
\multirow{3}{*}{Model}    & \multirow{3}{*}{Overall}          & \multicolumn{4}{c}{Diverse Q\&A}      & \multicolumn{2}{c}{Argument shifts} & \multirow{3}{*}{External} &    \multirow{3}{*}{\begin{tabular}[c]{@{}c@{}}Intent\\shifts\end{tabular}}     \\
\cmidrule(lr){3-6} \cmidrule(lr){7-8}
                                    &         & sQsA    & mQmA    & mQsA    & sQmA    & SO      & mSv     &         &         & \\ 
\midrule
GPT-4o (Prompt) & 72.00 & 72.52 & 75.63 & 73.00 & 74.05 & 72.98 & 61.98 & 69.64 & 76.22 \\
Claude3.5-sonnet (Prompt) & 72.02 & 76.62 & 75.52 & 73.34 & 73.94 & 76.49 & 69.68 & 71.83 & 58.74 \\
Qwen2.5-72B-Instruct (Prompt) & 68.12 & 64.59 & 67.82 & 63.90 & 67.40 & 65.40 & 64.98 & 61.12 & 89.79 \\
Llama-3.1-70B-Instruct (Prompt) & 70.49 & 75.05 & 73.20 & 70.74 & 73.52 & 70.25 & 68.87 & 69.29 & 63.02 \\
Qwen2.5-7B-Instruct (Prompt) & 57.78 & 60.08 & 58.12 & 54.40 & 57.70 & 54.21 & 51.85 & 53.87 & 72.04 \\
Llama-3.1-8B-Instruct (Prompt) & 51.49 & 51.53 & 57.70 & 48.59 & 56.22 & 48.84 & 48.30 & 50.14 & 50.63 \\
Ministral-8B-Instruct (Prompt) & 35.39 & 43.19 & 40.29 & 32.94 & 40.18 & 40.29 & 36.47 & 35.00 & 14.75 \\
Hammer2.1-7b (FC) & 61.31 & 62.26 & 60.86 & 52.63 & 60.65 & 57.85 & 52.81 & 48.73 & 94.71 \\
xLAM-7b-fc-r (FC) & 48.43 & 58.03 & 52.53 & 46.00 & 52.63 & 48.37 & 43.23 & 48.30 & 38.34 \\
ToolACE-8B (FC) & 48.19 & 54.54 & 52.63 & 38.55 & 51.68 & 50.25 & 45.74 & 49.15 & 42.98 \\
\bottomrule
\end{tabular}
\caption{The evaluation (Acc.\%) on \hb for different multi-turn data types. It is a snippet from \Cref{table:Overall}.}
\label{table:Multi-turn}
\end{table*}

\begin{table*}[htb]
\centering
\scriptsize
\setlength\tabcolsep{4pt}
\resizebox{1\textwidth}{!}{
\begin{tabular}{cccccccc}
\toprule
\multirow{3}{*}{Model}              & \multicolumn{6}{c}{w/o (PR) $|$  w/ ($\triangle$ PR)}                                                                                              & \multirow{3}{*}{IS} \\ \cmidrule{2-7}
                                    & \multicolumn{3}{c}{Diverse Q\&A}                                   & \multicolumn{2}{c}{Argument shifts}         & \multirow{2}{*}{External} &                     \\ \cmidrule(lr){2-4} \cmidrule(lr){5-6}
                                    & mQmA                 & mQsA                 & sQmA                 & SO                   & mSv                  &                           &                     \\ \midrule
GPT-4o (Prompt)                     & 66.03\% $|$ +7.189\% & 66.03\% $|$ +2.487\% & 66.03\% $|$ +4.763\% & 67.90\% $|$ +0.745\% & 68.47\% $|$ -12.01\% & 74.36\% $|$ -8.023\%      & 76.22\%             \\
Claude3.5-sonnet (Prompt)           & 70.64\% $|$ +3.101\% & 70.64\% $|$ -1.712\% & 70.64\% $|$ +1.977\% & 71.71\% $|$ +0.831\% & 71.99\% $|$ -8.048\% & 76.60\% $|$ -7.976\%      & 58.74\%             \\
Qwen2.5-72B-Instruct (Prompt)       & 56.28\% $|$ +8.470\% & 56.28\% $|$ +0.652\% & 56.28\% $|$ +7.913\% & 61.11\% $|$ -0.862\% & 63.15\% $|$ -5.797\% & 65.94\% $|$ -6.352\%      & 89.79\%             \\
Llama-3.1-70B-Instruct (Prompt)     & 64.72\% $|$ +2.704\% & 64.72\% $|$ -3.093\% & 64.72\% $|$ +3.282\% & 65.34\% $|$ -3.333\% & 65.70\% $|$ -7.057\% & 72.55\% $|$ -9.076\%      & 63.02\%             \\
Qwen2.5-7B-Instruct (Prompt)        & 48.90\% $|$ +4.159\% & 48.90\% $|$ -5.280\% & 48.90\% $|$ +4.138\% & 50.66\% $|$ -4.878\% & 51.71\% $|$ -9.961\% & 58.13\% $|$ -8.093\%      & 72.04\%             \\
Llama-3.1-8B-Instruct (Prompt)      & 38.43\% $|$ +12.72\% & 38.43\% $|$ -0.012\% & 38.43\% $|$ +11.92\% & 41.54\% $|$ -3.546\% & 42.76\% $|$ -7.786\% & 47.80\% $|$ -4.244\%      & 50.63\%             \\
Ministral-8B-Instruct-2410 (Prompt) & 26.78\% $|$ +1.805\% & 26.78\% $|$ -7.762\% & 26.78\% $|$ +1.589\% & 27.76\% $|$ -1.554\% & 28.17\% $|$ -6.172\% & 31.34\% $|$ -6.280\%      & 14.75\%             \\
Hammer2.1-7b (FC)                   & 52.37\% $|$ +5.194\% & 52.37\% $|$ -10.87\% & 52.37\% $|$ +4.806\% & 53.87\% $|$ -4.035\% & 54.76\% $|$ -10.89\% & 60.62\% $|$ -14.30\%      & 94.71\%             \\
xLAM-7b-fc-r (FC)                   & 39.28\% $|$ +1.457\% & 39.28\% $|$ -8.921\% & 39.28\% $|$ +1.793\% & 39.01\% $|$ -4.568\% & 39.14\% $|$ -10.05\% & 49.07\% $|$ -9.311\%      & 38.34\%             \\
ToolACE-8B (FC)                     & 42.55\% $|$ +3.843\% & 42.55\% $|$ -16.68\% & 42.55\% $|$ +3.554\% & 44.08\% $|$ -4.010\% & 44.81\% $|$ -9.745\% & 51.12\% $|$ -7.510\%      & 42.98\%             \\ \bottomrule
\end{tabular}
}
\caption{Multi-turn fine-grained evaluation across different data types for all snapshots. Each cell consists of two items: the absolute value of the baseline sQsA dataset and the change ($\triangle$) in the metric after encountering the corresponding situations. So that we can more clearly observe the impacts of these situations.}
\label{table:MT all snapshot}
\end{table*}

\begin{table*}[htb]
\centering
\scriptsize
\setlength\tabcolsep{4pt}
\resizebox{1.\textwidth}{!}{
\begin{tabular}{ccccccc}
\toprule
\multirow{3}{*}{Model}              & \multicolumn{6}{c}{w/o $|$ w/ ($\triangle$)}                                                                                                  \\ \cmidrule{2-7} 
                                    & \multicolumn{3}{c}{SO}                                             & \multicolumn{3}{c}{External}                                       \\ \cmidrule(lr){2-4} \cmidrule(lr){5-7} 
                                    & PHR               & PMR               & PR                   & PHR               & PMR               & PR                   \\ \midrule
GPT-4o (Prompt)                     & 7.555\% $|$ +0.501\% & 8.251\% $|$ -2.508\% & 68.86\% $|$ -0.996\% & 5.492\% $|$ +0.477\% & 5.492\% $|$ +18.76\% & 75.06\% $|$ -18.91\% \\
Claude3.5-sonnet (Prompt)           & 7.594\% $|$ +0.801\% & 6.845\% $|$ -3.348\% & 74.43\% $|$ -0.034\% & 4.690\% $|$ +2.987\% & 7.129\% $|$ +7.851\% & 75.54\% $|$ -15.61\% \\
Qwen2.5-72B-Instruct (Prompt)       & 11.88\% $|$ +2.471\% & 6.804\% $|$ -1.116\% & 66.32\% $|$ -4.742\% & 9.716\% $|$ +6.643\% & 8.097\% $|$ +6.217\% & 66.83\% $|$ -13.97\% \\
Llama-3.1-70B-Instruct (Prompt)     & 11.76\% $|$ +4.239\% & 2.023\% $|$ -0.858\% & 74.63\% $|$ -8.213\% & 8.302\% $|$ +5.055\% & 1.291\% $|$ +2.419\% & 78.94\% $|$ -15.71\% \\
Qwen2.5-7B-Instruct (Prompt)        & 16.83\% $|$ +7.338\% & 6.052\% $|$ -2.569\% & 60.75\% $|$ -14.70\% & 13.71\% $|$ +12.29\% & 4.219\% $|$ +3.391\% & 62.89\% $|$ -19.35\% \\
Llama-3.1-8B-Instruct (Prompt)      & 16.57\% $|$ +6.999\% & 23.39\% $|$ -5.054\% & 51.68\% $|$ -9.725\% & 12.28\% $|$ +9.829\% & 23.81\% $|$ -6.427\% & 54.69\% $|$ -10.10\% \\
Ministral-8B-Instruct-2410 (Prompt) & 21.67\% $|$ +7.864\% & 6.454\% $|$ -3.440\% & 50.72\% $|$ -9.175\% & 17.56\% $|$ +9.367\% & 4.683\% $|$ +14.28\% & 53.16\% $|$ -23.74\% \\
Hammer2.1-7b (FC)                   & 14.13\% $|$ +7.616\% & 8.469\% $|$ -3.306\% & 61.58\% $|$ -11.20\% & 11.52\% $|$ +4.699\% & 5.555\% $|$ +29.96\% & 64.72\% $|$ -31.59\% \\
xLAM-7b-fc-r (FC)                   & 19.47\% $|$ +10.16\% & 8.571\% $|$ +0.313\% & 62.33\% $|$ -19.62\% & 14.25\% $|$ +7.414\% & 6.653\% $|$ +14.06\% & 66.73\% $|$ -23.74\% \\
ToolACE-8B (FC)                     & 27.29\% $|$ +10.78\% & 4.461\% $|$ -2.101\% & 53.67\% $|$ -12.50\% & 23.38\% $|$ +11.29\% & 3.225\% $|$ +5.040\% & 58.06\% $|$ -19.42\% \\ \bottomrule
\end{tabular}
}
\caption{Evaluations for snapshots at the moment
of slot overriding (SO) and answering with pronouns (External). Each cell in the table includes two items: the baseline absolute metrics under the sQsA dataset, and the changes ($\triangle$) in metrics after considering SO/External.}
\label{table:MT SO EK}
\end{table*}

\end{document}